\documentclass[11pt]{article}
\usepackage[table, dvipsnames]{xcolor}

\usepackage[final]{acl}

\usepackage{times}
\usepackage{latexsym}

\usepackage[T1]{fontenc}

\usepackage[utf8]{inputenc}

\usepackage{microtype}

\usepackage{inconsolata}

\usepackage{graphicx}

\usepackage{hyperref}
\usepackage{url}
\usepackage{booktabs}

\usepackage{lineno}

\usepackage{fontawesome}
\usepackage{siunitx}
\usepackage{xspace}
\usepackage[inline]{enumitem}
\usepackage[multiple]{footmisc}
\usepackage{mleftright}
\usepackage{cuted,tcolorbox}
\usepackage{xurl}
\usepackage{multirow}
\usepackage{listings}

\usepackage[noend]{algpseudocode}
\usepackage{algorithm}
\usepackage{amsfonts}
\usepackage{amsmath}
\usepackage{amssymb}
\usepackage{bm}
\usepackage{amsthm}
\usepackage{mathtools}
\usepackage{relsize}
\usepackage{hyphenat}
\usepackage{graphicx}
\usepackage{makecell}
\usepackage{pifont}
\usepackage{wrapfig}
\usepackage{subfigure}
\usepackage{CJKutf8}

\DeclarePairedDelimiterX{\infdivx}[2]{(}{)}{%
  #1\;\delimsize\|\;#2%
}

\definecolor{cbars0}{HTML}{FF5E5E}
\definecolor{cbars1}{HTML}{FABE28}
\definecolor{cbars2}{HTML}{0dbae7}
\definecolor{cbars3}{HTML}{465cee}
\definecolor{cbars4}{HTML}{d3eff6}
\definecolor{cbars5}{HTML}{2B3E51}
\definecolor{cbars6}{HTML}{76D7EA}
\definecolor{cbars_border}{HTML}{696969}

\definecolor{c1}{HTML}{E85642}

\definecolor{darkblue}{rgb}{0, 0, 0.5}
\hypersetup{colorlinks=true, citecolor=darkblue, linkcolor=darkblue, urlcolor=darkblue}

\title{On the Consistency of Multilingual Context Utilization \\ in Retrieval-Augmented Generation}

\author{
  Jirui Qi$^1$ ~\;~ Raquel Fern\'{a}ndez$^2$ ~\;~ Arianna Bisazza$^1$\vspace{2pt}\\
  $^1$Center for Language and Cognition (CLCG), University of Groningen \\
  $^2$Institute for Logic, Language and Computation (ILLC), University of Amsterdam \\
  \texttt{\{j.qi, a.bisazza\}@rug.nl, raquel.fernandez@uva.nl}\\
  }

\begin{document}
\maketitle
\begin{abstract}
Retrieval-augmented generation (RAG) with large language models (LLMs) has demonstrated strong performance in multilingual question-answering (QA) tasks by leveraging relevant passages retrieved from corpora. In multilingual RAG (mRAG), the retrieved passages can be written in languages other than that of the query entered by the user, 
making it challenging for LLMs to effectively utilize the provided information.
Recent research suggests that retrieving passages from multilingual corpora can improve RAG performance, particularly for low-resource languages. 
However, the extent to which LLMs can leverage different kinds of multilingual contexts to generate accurate answers, \textit{independently from retrieval quality}, remains understudied.
In this paper, we conduct an extensive assessment of LLMs' ability to (i) make consistent use of a relevant passage regardless of its language, 
(ii) respond in the expected language, and 
(iii) focus on the relevant passage even when multiple `distracting' passages in different languages are provided in the context.
Our experiments with four LLMs across three QA datasets covering 48 languages reveal a surprising ability of LLMs to extract relevant information from passages in a different language than the query, but a much weaker ability to produce a full answer in the correct language.
Our analysis, based on both accuracy and feature attribution techniques, further shows that distracting passages negatively impact answer quality regardless of their language. However, distractors in the query language exert a slightly stronger influence.
Taken together, our findings deepen the understanding of
how LLMs utilize context in mRAG systems, providing directions for future improvements. \footnote{All codes and data released at \url{https://github.com/Betswish/mRAG-Context-Consistency}.} 
\end{abstract}

\section{Introduction}
Retrieval-augmented generation has shown strong results in multilingual question-answering (QA) tasks \citep{chirkova-etal-2024-retrieval, thakur2024mirage}. Given a query in the user language, informative passages are retrieved from a reference corpus and provided jointly with the query, promoting the large language model (LLM) to generate more precise responses \citep{lewis2020retrieval, asai2021one}. 
In multilingual RAG (mRAG), retrieval can be performed either monolingually or cross-lingually.
In the former, retrieval is performed only over passages in the same language as the query \citep{asai-etal-2023-retrieval, gao2023retrieval, fan2024survey}, which can be successful for high-resource languages. However, this approach is marginally useful, or even harmful, when the question is posed in a low-resource language, since relevant information is likely to be available only in different languages \citep{muller-etal-2023-evaluating}.
In addition, for questions regarding a specific geographical region or culture, essential information may be present only in corpora of the languages spoken in that region. 
To address this issue, cross-lingual retrieval attempts to extract useful information 
simultaneously from multiple languages
\citep{asai2021one, li-etal-2024-bordirlines}, leading to visible gains in low-resource languages \citep{chirkova-etal-2024-retrieval}


\begin{figure*}[!t]
    \centering
    \includegraphics[width=0.99\linewidth]{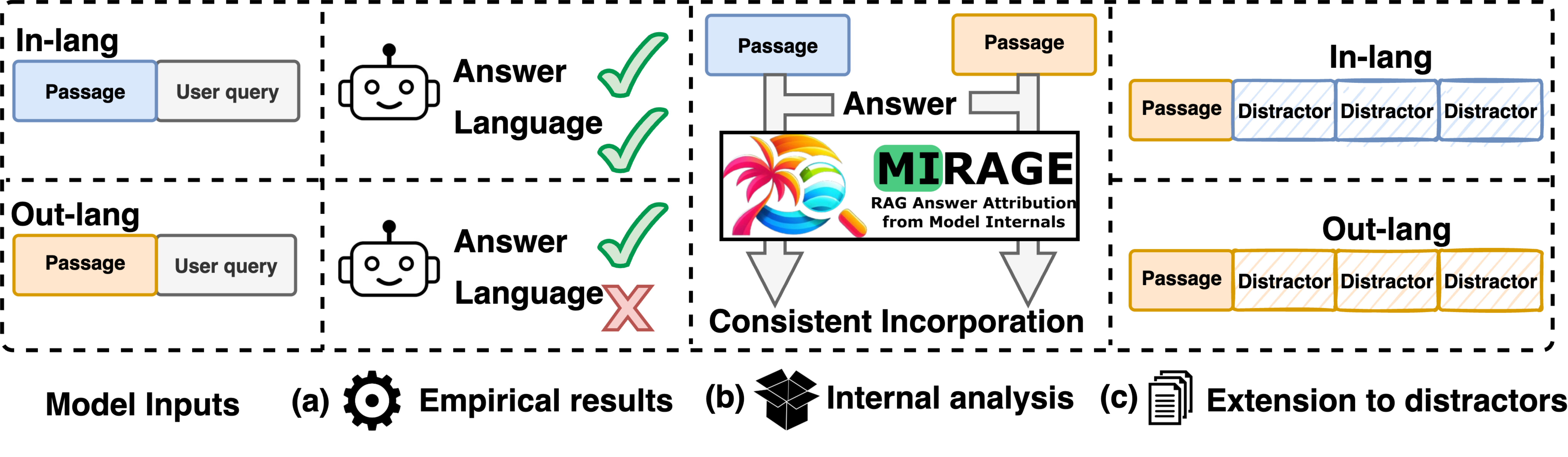}
    \caption{Illustration of the contributions and proposed assessment frameworks of this paper.
    }
    \label{fig:background}
\end{figure*}

Evaluating RAG pipelines is notoriously difficult due to the open-endedness of the retrieval task, and to the complex interactions of retrieval quality with model understanding and generation abilities.
On top of this, multilinguality adds another layer of complexity.
Ideally, retrieved passages should be equally useful when the same question is posed in different languages. 
Besides, LLM-generated answers should be consistently correct across languages so that users with different language backgrounds enjoy a similar experience. 
However, despite the reported accuracy improvements, the abilities of LLMs to exploit cross-lingually retrieved contexts in mRAG remain poorly understood. 

In this paper, we conduct an in-depth assessment of these abilities, using standard accuracy evaluation as well as feature attribution analysis.
Unlike recent mRAG evaluations \citep{chirkova-etal-2024-retrieval, park2025investigating}, which test the LLM performance for each language in the entire RAG pipeline (i.e., retrieval + generation), we disentangle these two factors and focus on the LLM's ability to exploit context independently from retrieval quality.
As shown in Figure \ref{fig:background}, our key contributions include: 
(a)~We evaluate how LLMs leverage retrieved passages in different languages in various multilingual QA tasks, revealing remarkably robust input understanding but much more brittle generation abilities. 
(b)~Besides the standard accuracy evaluation, we apply a recently proposed RAG answer attribution method based on model internals \citep{qi-etal-2024-model} to confirm that LLMs consistently incorporate retrieved content from various languages, providing insights from an interpretability perspective. 
(c)~We consider both single-passage and multi-passage mRAG setups and examine how distracting passages in different languages affect model performance, shedding light on the complex interplay between relevance and content of the retrieved passages.  
Taken together, our results deepen our understanding of how LLMs utilize context in mRAG systems and reveal important areas for future improvements.

\section{Related Work}
\subsection{Retrieval Strategies for mRAG}
Retrieval is a key component of mRAG, which can be performed in at least two ways: monolingually (in-language) or cross-lingually. 
\citealt{chirkova-etal-2024-retrieval} investigated mRAG systems across 13 languages, highlighting the limited gains of in-language retrieval in their setup. 
\citealt{nie-etal-2023-cross} proposed the Prompts Augmented by Retrieval Crosslingually (PARC) pipeline, which augments contexts with semantically similar sentences retrieved from high-resource languages to enhance zero-shot performance in low-resource languages. \citealt{gao-etal-2022-retrieval} introduced a retrieval-augmented method for multilingual keyphrase generation, leveraging keyphrase annotations in English to aid keyphrase generation in low-resource languages through cross-lingual dense passage retrieval.

\begin{table*}[!t]
\centering
\resizebox{\textwidth}{!}{
\begin{tabular}{c|c@{\ \ }c|cc c@{\ \ }c@{\ \ }c c}
\toprule
\multirow{2}{*}{\bf Dataset} & {\bf QA Task} & \multirow{2}{*}{\bf \# Languages} & \multirow{2}{*}{\bf \# Queries} & {\bf \# Queries}  & \multicolumn{3}{c}{\bf Parallel?} & {\bf Answer}\\
& {\bf Type} & & & {\bf(w/ Gold Pass.)} & Query & Answer & Gold Pass. & {\bf Format} \\
\midrule
XQUAD & Extractive & 12 & 1190 & 1190 & \color{LimeGreen}\ding{51} & \color{LimeGreen}\ding{51} & \color{LimeGreen}\ding{51} & Text\\
MKQA & Open Domain & 24 & 6758 & 5951 & \color{LimeGreen}\ding{51} & \color{LimeGreen}\ding{51} & \color{WildStrawberry}\ding{55} & Text \\
GMMLU & Multi-Choice & 42 & 14042 & 4136 & \color{LimeGreen}\ding{51} & \color{LimeGreen}\ding{51} & \color{WildStrawberry}\ding{55} &  A/B/C/D \\
\bottomrule
\end{tabular}
}
\caption{Overall dataset statistics. {\bf \# Queries (w/ Gold Pass.)} refers to the number of queries with at least one gold passage in any language, which is the subset used for our experiments (cf. Section~\ref{sec:retrieval_and_filtering}).}
\label{tab:dataset_statistic}
\end{table*}

\subsection{Consistency in Multilingual LLMs}
Ensuring model consistency across languages is a key objective for multilingual LLMs. A series of recent works has focused on the consistency of factual knowledge encoded in the weights of multilingually pre-trained LLMs \citep{fierro-sogaard-2022-factual,weber-etal-2023-mind, qi-etal-2023-cross, genbench-2023-genbench}. 
Other work has focused on the consistency of domain-specific QA by assessing whether the questions asked by a certain group of people \citep{schlicht2025llms} or about domain-specific knowledge \citep{yin-etal-2022-geomlama, Li2024MultilingualRA} can be correctly answered by LLMs regardless of the query language.
Very recently and concurrently with our work, research interest has also risen around the consistency of mRAG pipelines \citep{wu2024not,sharma2024faux,park2025investigating}. 

\subsection{Context Utilization in mRAG}
Although some studies \citep{asai2021one, nie-etal-2023-cross, stap-monz-2023-multilingual, chirkova-etal-2024-retrieval} have demonstrated that cross-lingual retrieval can significantly enhance mRAG answer accuracy, the extent to which LLMs can utilize multilingual contexts consistently remains poorly understood, motivating the present work. 
The concept of \textit{context utilization} is also not always clearly defined. Recent and concurrent studies \citep{wu2024not,sharma2024faux,park2025investigating} 
use performance of the complete mRAG pipeline 
to study context utilization 
and find that models tend to prefer passages in the query language or Latin scripts. 
In this paper, we further
distinguish between \textit{input understanding} and \textit{decoding capability} as key abilities of an mRAG generator, and disentangle them through our experiments, while strictly controlling for retrieval quality. 

\section{Experimental Setup}

Consider a multilingual QA setup where $q^\ell$ is a query in language $\ell$ and
$a^\ell$ is the gold answer in the same language.
For each query, a set of relevant passages $P_q=\{p_1, \dots, p_n\}$ in multiple languages is retrieved from a reference multilingual corpus. 
A relevant passage ($p \in P_q$) is considered gold $\widehat{p}$ if it includes the necessary information to answer $q^\ell$ correctly, or non-gold (`distracting') $\overline{p}$ otherwise. 
To perform mRAG, a subset of relevant passages $C_q \subset P_q$ is selected and provided as extra context to the LLM along with query $q^\ell$.
In an ideal mRAG setting, the model should answer more accurately when provided with $C$ but it should also be \textit{agnostic to the languages} in which the passages $p \in C$ are provided, in terms of both answer accuracy and feature attribution results. 
Following \citealp{muller-etal-2023-evaluating}, 
we use the term `\textbf{in-language}' for the same language as the user query language, and `\textbf{out-language}' for different languages than the user language. 

Given this setup, we study LLMs' ability to handle multilingual context in different retrieval scenarios, which we simulate by varying (i) the number of gold and non-gold (`distracting') passages provided in $C$, and (ii) the languages of those passages.

\subsection{Datasets} 
Question answering datasets can differ across many dimensions. We choose three multilingual QA benchmarks to cover a diverse set of languages, three different types of QA, and different levels of parallelism (see Table~\ref{tab:dataset_statistic}) allowing us to isolate different aspects of mRAG in our evaluation. 

\textbf{XQUAD} \citep{artetxe-etal-2020-cross} is an extension of the extractive English QA dataset SQUAD \citep{rajpurkar-etal-2016-squad}, which contains 1190 questions, each provided with a single relevant passage and a gold answer, all translated into 12 languages. While not being originally designed for RAG evaluation, this dataset is the only one allowing us to assess LLMs' abilities to use \textit{the exact same information} provided in different languages, simulating an impossible scenario where retrieval works perfectly in all languages.
%
\textbf{MKQA} \citep{longpre-etal-2021-mkqa} is an open domain QA dataset covering 10,000 questions across 24 languages
derived from Natural Questions \citep{kwiatkowski-etal-2019-natural}. Removing the questions without any gold answers provided, we work on a total of 6758 paralleled questions in this paper.
\textbf{Global-MMLU} or GMMLU \citep{singh2024global} is a large multilingual extension of MMLU \citep{Hendrycks2020MeasuringMM} obtained by translating the English instances into 41 languages. Like MMLU, it contains 14042 multi-choice questions that are used to test LLMs' understanding capability across a range of subjects, like social sciences or medical questions. Each question is provided with four options to choose from. 
Question examples for all datasets are given in Appendix~\ref{sec:dataset_examples}.

\subsection{Retrieval and Filtering}
\label{sec:retrieval_and_filtering}
\textbf{XQUAD} includes a single gold passage for each query, which we can provide to the model without performing any retrieval ($C_q=P_q=\{\widehat{p}\}$).

As for \textbf{MKQA} and \textbf{GMMLU}, we retrieve passages from Multilingual Wikipedia Corpora\footnote{\url{https://huggingface.co/datasets/wikimedia/wikipedia}} using the Cohere Embed Multilingual V3 retriever\footnote{\url{https://huggingface.co/Cohere/Cohere-embed-multilingual-v3.0}}, a strong performing multilingual embedding model with balanced language coverage \citep{cohere2023embedv3}. 
Unlike previous work \citep{asai2021one, muller-etal-2023-evaluating, chirkova-etal-2024-retrieval} where the number of studied languages was at most 13, our evaluation covers twice or more languages, making it unfeasible to perform a full cross-lingual retrieval for each query language. As an approximation, we construct the set of relevant passages $P_q$ by performing in-language retrieval for the $L$ parallel versions of $q$ in each language and taking the union of the top-30 ranked passages in each language:
$P_q = \bigcup_{\ell=1}^{L} P_{q^\ell}$. 

Then, we tag the gold passages in $P_q$ based on whether they contain the gold answer as a substring, following previous work \citep{liu-etal-2024-lost, liu2024likelihood}. 
In our experiments, we only consider queries for which $P_q$ contains at least one gold passage in any of the studied languages, see resulting \mbox{\# Queries (w/ Gold Pass.)} in Table~\ref{tab:dataset_statistic}.
%
While it may be possible to expand this subset by retrieving more than 30 top passages or by improving retriever quality \citep{chirkova-etal-2024-retrieval}, 
we believe our setup is appropriate to study LLMs' ability to use a variety of multilingual context types that are representative of competitive cross-lingual retrieval results.\footnote{Although a large portion of GMMLU queries are filtered out, we argue that the remaining 4136 queries are numerous enough to ensure a robust evaluation. We also verify the diversity of this subset and find a total of 55 covered subjects. See Appendix \ref{sec:covered_subjects} for details on the question subjects and categories.} 

Detailed statistics on the amount of in-language and out-language gold passages for all queries are shown in Appendix \ref{sec:full_statistics}.
As expected, the situation is particularly serious for queries posed in low-resource languages, where only out-language gold passages are available for most of the queries (e.g., 88\% in Khmer MKQA and 91\% in Yoruba GMMLU), highlighting the importance of ensuring mRAG quality across many languages. 

\subsection{Evaluation Metrics}
\label{sec:metrics}
For \textbf{XQUAD} and \textbf{MKQA}, we follow previous work \citep{asai2021one} and score answers by strict lexical matching, that is, 1 if the entire gold answer string $a^\ell$ is a substring of the model response $\mathcal{M}(q^\ell)$, or 0 otherwise. 
Since models in mRAG setups often generate the correct answer in the wrong passage language \citep{chirkova-etal-2024-retrieval, zhang-etal-2024-respond}, we also measure the proportion of model answers that contain a gold answer in language $\ell'$  ($a^{\ell'}, \ell'\neq\ell$). 
\footnote{
Since we focus on the language of model responses and outright cross-language generation (i.e., whether the gold answer appears in a different language) where small orthographic variants can be decisive, particularly for phonologically similar languages, we do not adopt the variant of the softer lexical metric \citep{chirkova-etal-2024-retrieval} (3-gram recall), which tolerates minor orthographic differences and could blur the distinctions.
}
Nevertheless, as exact matching could be overly strict, we further adopt two complementary metrics (BERTScore and GPT-4.1-nano) on XQUAD. Similar results are observed, providing more insights and enhancing the robustness of our analysis. See Appendix \ref{app:extensive_eval} for more details.

\textbf{GMMLU} is instead designed as a multi-choice task, thus, accuracy can be simply evaluated by checking if the LLM outputs the correct option letter (A/B/C/D). 
To study the impact of answer generation from that of passage understanding across languages, we also use GMMLU as an open QA task by providing the query without any answer options, and adopting again lexical matching for evaluation.
We refer to the original dataset as \textbf{GMMLU-Choice}, and the no-options one as \textbf{GMMLU-Open}.

\subsection{Models}
We evaluate four top-performing multilingual LLMs belonging to different model families, which have been used in recent mRAG evaluations \citep{wang2024retrieval, thakur2024mirage}, namely: Aya-Expanse-8B \citep{dang2024ayaexpansecombiningresearch}, 
Llama-3.2-3B-Instruct \citep{dubey2024llama}, 
Gemma-2-9B-it \citep{gemma_2024}, and 
Qwen2.5-7B-Instruct \citep{qwen2}. Although these models do not officially support some of our studied languages, evidence has shown that LLMs can generalize successfully to unseen languages due to the leak of training data or shared representations \citep{qi-etal-2023-cross, budnikov2024generalization, lu2024every}, which we also observed in preliminary experiments.

\begin{figure*}[!t]
    \centering
    \includegraphics[width=1\linewidth]{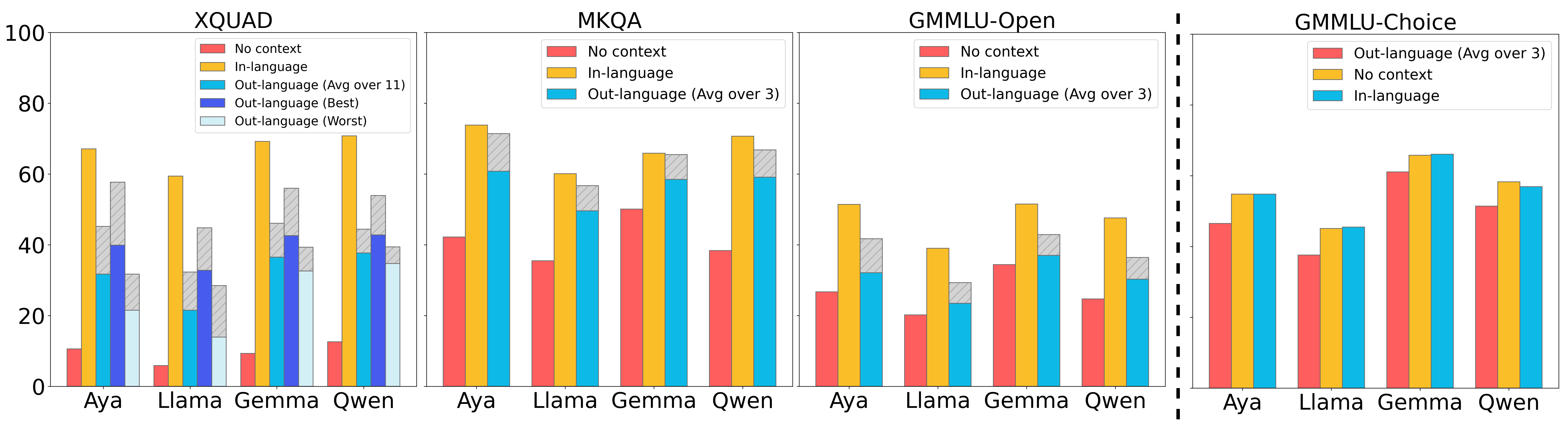}
    \caption{Performance on XQUAD, MKQA, GMMLU-Open, and GMMLU-Choice, where the LLMs are provided with no retrieved passage or one gold passage in either in-language or out-language. The shading on the bars represents the ratio of questions that can be correctly answered but in the wrong passage language, which does not apply to GMMLU-Choice since the evaluation on it is not affected by the generation language. }
    \label{fig:single-doc}
\end{figure*}

\section{Single-Passage mRAG}
\label{sec:single-passage}
We start from a simple scenario where, for each query $q^\ell$, only one gold passage is provided to the model either in the query language (in-language; $C=\{\widehat{p}^\ell\}$) or in a different language  (out-language; $C=\{\widehat{p}^{\ell'}\}, \ell' \neq \ell$). 
As a baseline, we calculate answer accuracy when no context is provided to the model ($C=\emptyset$). 

In XQUAD, where gold passages are translated into 12 languages, we iterate over the 11 out-language passage versions for each query and report the average accuracy. We also report accuracy for the passage language that yielded the best (or worst) answer accuracy overall for each query language. 
By contrast, the gold passages in MKQA and GMMLU are retrieved from a Wikipedia corpus as explained in Section~\ref{sec:retrieval_and_filtering}, and are not parallel across languages. 
As a solution, for each query $q^\ell$, we randomly sample 3 different out-language passages from $P_q$ and report accuracy averaged over the 3 single-passage answers.
%
To maximize the chances of obtaining a model response in the query language $\ell$, we explicitly mention $\ell$ in the instruction, which is itself translated into $\ell$, following \citealp{chirkova-etal-2024-retrieval, zhang-etal-2024-respond}.
The detailed prompts are listed in Appendix \ref{sec:prompts}. 



\subsection{Accuracy Results}
\label{sec:accuracy_results}


Results averaged across all query languages are given in Figure \ref{fig:single-doc}, while the full language-specific results are given in Appendix \ref{sec:full_results}. 

\paragraph{Results on XQUAD}
We recall that XQUAD is a distinct dataset, originally developed to evaluate extractive QA, rather than open-domain RAG systems. 
Nevertheless, it is the only dataset where the exact same gold passage is available in different languages, allowing us to isolate the effect of a passage's language from that of its content.
%
As shown in Figure \ref{fig:single-doc}, providing the gold passage in any language strongly improves answer accuracy compared to the \textcolor{cbars0}{no-context} baseline, which is likely due in part to the extractive nature of QA in this dataset. 
Looking at the passage language, however, we find that \textcolor{cbars1}{in-language} passages yield considerably higher accuracy than all \textcolor{cbars2}{out-language} settings, including out-language (Best).
Moreover, a notable portion of questions are answered \textcolor{cbars_border}{correctly but in the wrong language} even though the models were explicitly prompted to answer in the target language, which is in line with previous findings \citep{wu2024not, chirkova-etal-2024-retrieval}.
Even when considering these cases, a visible gap remains between in-language and out-language accuracy across the board on XQUAD. We further analyze this gap through manual error analysis and find that missed matches are often due to the use of synonyms or slight paraphrases of the gold answer, or --in the case of languages with different scripts-- to transliteration variations \citep{knight-graehl-1997-machine}. See Appendix~\ref{sec:error_analysis} for more details. 

\begin{figure}[!t]
    \centering
    \includegraphics[width=.9\linewidth]{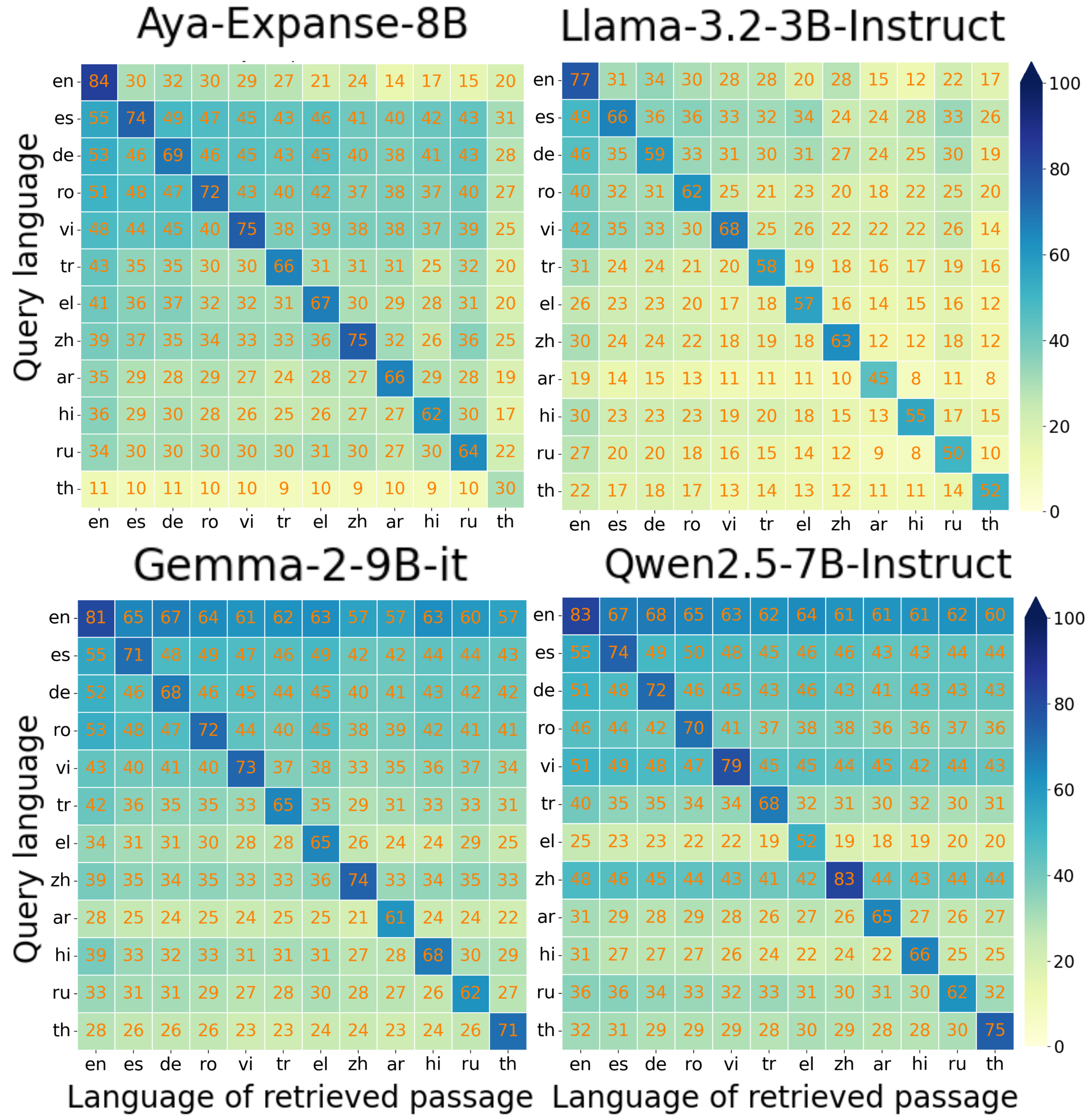}
    \caption{Answer accuracy (\%) on XQUAD among different query-passage language combinations. Only model answers in the correct (i.e., query) language are considered as correct.
    }
    \label{fig:heatmap-strict}
\end{figure}



Figure~\ref{fig:heatmap-strict} gives a detailed view of how answer accuracy varies with the language of the provided gold passage.\footnote{Here we only consider answers in the correct language, see Appendix \ref{sec:full_results} for language-specific accuracies when considering the wrong generation language.} As expected, the highest accuracy is always achieved when the retrieved passage is in the same language as the query. 
Concurrent work \citep{sharma2024faux,park2025investigating} suggested that models may prefer passage languages that use the same script as the query language, based on a few languages.
Because script similarity is a very coarse-grained measure of language similarity that is not informative for many of our language pairs, we turn to finer-grained measures that were previously shown to strongly correlate with cross-lingual consistency of model answers in non-RAG setups \citep{qi-etal-2023-cross}. In particular, we adopt \textit{subword vocabulary overlap} computed on a reference parallel corpus\footnote{Following \citealp{qi-etal-2023-cross}, we extract the vocabularies from FLORES-200 \citep{costa2022no}, a strictly parallel corpus covering 200 languages, and measure their pairwise overlap via Jaccard index \citep{jaccard1912distribution}.} 
as this was shown to correlate better with response consistency than various typological similarity measures.
We compute Pearson and Spearman correlations
between subword overlap and answer accuracy for each query language (excluding the case where query and passage are in the same language), however all correlations are low and not statistically significant. Looking back at Figure~\ref{fig:heatmap-strict}, we notice that shading (i.e., answer accuracy) is relatively consistent within each row, especially on Gemma and Qwen, more so than within each column. In other words, the query language is much more predictive of accuracy than the passage language, suggesting that generating in the target language is the major bottleneck in our setup, which could dominate, if not hide, the effect of similarity with the passage language. 

Additionally, we also investigate if advanced prompts with multi-step instructions (refer to Appendix \ref{app:advanced_prompt}) or larger models (open-source Gemma3-27B-IT and closed-source GPT-5-nano estimated at 8-18B parameters\footnote{\url{https://www.r-bloggers.com/2025/08/how-many-parameters-does-gpt-5-have/}}; Appendix \ref{app:larger_model}) can mitigate the language mismatch issue in model answers. However, the problem persists, further reinforcing our finding that multilingual RAG systems face an inherent decoding limitation.
Interestingly, we observe that when fed with passages in Thai, which is not officially supported by Aya-Expanse-8B, the model always outperforms the baseline where no context is provided for queries in each language (cf. No-context accuracies in Table~\ref{tab:full_XQUAD_1}). This suggests that even though the passages are written in a language that is unseen in the pre-training phase, LLMs may be able to utilize them. 

\paragraph{Results on MKQA}
Moving to a more realistic RAG dataset, but without parallel passages, we find a similar trend (Figure~\ref{fig:single-doc}) where in-language gold passages outperform out-language ones, however the gap is much smaller than in XQUAD and almost disappears when also considering the portion of questions that are answered correctly in the wrong language. 
These results suggest that the passage language is not a key factor blocking LLMs from understanding and utilizing the context in MKQA.


\paragraph{Results on GMMLU}
Accuracy results on GMMLU-Open (Figure~\ref{fig:single-doc}) are in line with the two previous datasets, with an in-/out-language gap falling halfway, that is smaller than in XQUAD but larger than in MKQA.
To further disentangle the impact of context understanding from that of target language generation, we compare these results with those of GMMLU-Choice, where the model only has to generate one of the four option letters (A/B/C/D) provided in the prompt.
Here, we find that in- and out-language passages yield extremely close accuracy, confirming that input understanding is not the real obstacle for high-quality mRAG.
Rather, the main barrier appears to lie on the side of generation, namely, whether models can formulate a proper response \textit{in the correct target language}. 




\subsection{Interpretability-based Assessment}
\label{sec:single-passage-mirage}

To further verify our findings that the passage language is not a barrier to LLMs' understanding capability of the multilingual retrieved passages, we adopt MIRAGE \citep{qi-etal-2024-model}, a model internal-based method for attributing model responses to the retrieved passages in RAG systems. Generally, it consists of two components: (1) CTI for detecting contextual sensitivity for the generated sentence and (2) CCI for attributing the detected sentences back to each retrieved passage.
Given the single-passage setup, 
in this section we only use the CTI module for evaluating the passage dependency of the model response. For each generated token, this module measures the shift in output probability distribution when no context vs. one passage is provided, measured by KL divergence \citep{kldiv}, while keeping the generated sentence prefix fixed. 
If at least one token is higher than an empirically set CTI threshold, the generated sentence is marked as sensitive to the context provided in the prompt. 

\begin{table}[!t]
\centering
\resizebox{0.99\linewidth}{!}{
\begin{tabular}{ccccccc}
\toprule
\multirow{2}{*}{\bf Dataset} 
        & \multicolumn{2}{c}{\bf AVG+1.0*SD} 
        & \multicolumn{2}{c}{\bf AVG+1.5*SD} 
        & \multicolumn{2}{c}{\bf AVG+2.0*SD} \\
\cmidrule(lr){2-3} \cmidrule(lr){4-5} \cmidrule(lr){6-7}
& In. & Out. & In. & Out. & In. & Out. \\
\midrule
XQUAD & 98 & 99 & 97 & 99 & 95 & 98 \\
MKQA  & 100 & 100 & 100 & 100 & 100 & 100 \\
\bottomrule
\end{tabular}
}
\caption{Percentage (\%) of context-sensitive responses when Aya is provided with in-language (In.) vs. out-language (Out.) gold passages, detected by MIRAGE under different CTI thresholds. }
\label{tab:mirage_single}
\end{table}

\begin{table*}[!t]
\centering
\begin{tabular}{lcccccccc}
\toprule
\multirow{2}{*}{\bf Setup} & \multicolumn{4}{c}{\bf MKQA} & \multicolumn{4}{c}{\bf GMMLU-Choice} \\
\cmidrule(lr){2-5} \cmidrule(lr){6-9}
& \textbf{Aya} & \textbf{Llama} & \textbf{Gemma} & \textbf{Qwen} & \textbf{Aya} & \textbf{Llama} & \textbf{Gemma} & \textbf{Qwen} \\
\midrule
No Ctx              & 41.4  & 34.9  & 49.6  & 37.7  & 46.6  & 37.8  & 61.1  & 51.4  \\
\midrule
1 Gold (in)         & 73.6  & 59.9  & 65.5  & 70.6  & 55.0  & 45.2  & 66.0  & 58.4  \\
\midrule
1 Gold (out) & 71.1 & 56.5 & 65.1 & 66.6 & 55.0 & 45.5 & 66.2 & 57.0 \\
\ + 3 Dist (in) & 47.0 & 39.4 & 53.7 & 49.6 & 50.8 & 41.1 & 65.1 & 54.5 \\
\ + 3 Dist (out) & 47.7 & 38.9 & 56.1 & 53.8 & 51.4 & 42.5 & 64.8 & 54.6 \\
\midrule 
3 Gold (out) & 77.7 & 65.3 & 75.3 & 75.9 & 56.6 & 47.7 & 69.1 & 59.5 \\
\ + 1 Dist (in) & 68.8 & 56.0 & 71.3 & 70.8 & 55.8 & 45.7 & 68.6 & 58.4 \\
\ + 1 Dist (out) & 69.6 & 57.8 & 72.9 & 72.9 & 56.2 & 46.6 & 68.5 & 58.6 \\
\bottomrule
\end{tabular}
\caption{Average answer accuracy (\%) without context (No Ctx), with a single in-language gold passage (1 Gold (in)), and multi-passage mRAG setups 
with varying numbers of in-language or out-language gold passages and distracting passages. Results are averaged over all query languages.}
\label{tab:multi_passage}
\end{table*}

We select Aya-Expanse-8B as the studied model and sample 500 instances separately from XQUAD and MKQA. 
Table \ref{tab:mirage_single} shows the results under different CTI thresholds.
We find that nearly all generated responses are tagged as context-sensitive by MIRAGE, even when setting a higher CTI threshold (${\rm avg}$ + 2 $\* {\rm std\_dev}$) than the one used in the original paper.
This confirms that the provided passage significantly drives models' predictions regardless of its language. 
 

In sum, the results in this section point to the fact that understanding passages in different languages and locating useful information within them is not the main obstacle towards high-quality mRAG, whereas generation abilities in several target languages remain a serious bottleneck.
In the next section, we study how models handle more realistic contexts consisting of multiple passages in different languages.

\section{Multi-Passage mRAG}
Real-world RAG settings are further complicated by the presence of multiple passages \mbox{$C_q = \{p_1, \dots, p_n\}$}, some of which may be related to the query but not functional to answering it correctly (i.e. `distracting' passages $\overline{p}$).
We investigate how the language of different passages in the context affects LLMs' ability to locate the right information, assuming this is included in at least one passage of the context.
In particular, we aim to assess model robustness in a challenging scenario where the important information is only provided in a different language than the query, along with several in-language distractors. 

For simplicity, we set the maximum number of passages to 4 and 
simulate two practical scenarios: 
(i) a weak retriever finds one out-language gold passage while the other three are distractors; 
(ii) a strong retriever finds three out-language gold passages while the remaining one is a distractor.
In both cases, we compare accuracies when the distractors are in-language vs. out-language. 
We conduct experiments on MKQA and GMMLU-Choice. XQUAD is excluded because it is an extractive QA dataset, unsuitable for multi-passage mRAG.


\subsection{Accuracy Results}


Table~\ref{tab:multi_passage} presents the results, including the no-context baseline and single in-/out-language gold passage results as computed in Section \ref{sec:single-passage}, to enable comparison (see Appendix \ref{sec:full_results} for full language-specific results). For this analysis, we also consider as valid the questions that were answered correctly but in the wrong language, as they also reflect a proper understanding of the context by the model.
Interestingly, models provided with 3 out-language gold passages achieve higher accuracy than when provided with a single in-language gold passage in the query language, emphasizing the potential of cross-lingual retrieval for mRAG.
As expected, the presence of distractors leads to lower accuracy. Notably, this is true for all models, datasets, and setups. However, the effect is considerably stronger in MKQA than in GMMLU-Choice, likely due to the stricter lexical-matching metric adopted for MKQA. 
We also verify that a higher proportion of distractors (3/4 vs. 1/4) is much more harmful for answer accuracy, which confirms the importance of having access to a high-quality cross-lingual retriever \citep{chirkova-etal-2024-retrieval}.
%
When comparing the drop between in-language distractors and out-language distractors, we find that in-language distractors have a larger impact in most cases, matching our hypothesis that this is a particularly challenging scenario for LLMs.
However, differences are small in many cases, indicating the language of the distractor is not a major issue for multi-passage mRAG. 






\begin{table}
\centering
\begin{tabular}{lcccc}
\toprule
\multirow{2}{*}{\bf AVG Dist.} & \multicolumn{2}{c}{\bf MKQA} & \multicolumn{2}{c}{\bf GMMLU} \\
\cmidrule(lr){2-3} \cmidrule(lr){4-5}
& In. & Out. & In. & Out. \\
\midrule
\rowcolor{black!10} \multicolumn{5}{c}{\bf 1 Gold (out) + 3 Distractors} \\
\midrule
\# Context. & 1.77 & 1.74 & 1.89 & 1.82 \\
\# Influent. & 0.94 & 0.86 & 1.13 & 1.07 \\
\midrule
\rowcolor{black!10} \multicolumn{5}{c}{\bf 3 Gold (out) + 1 Distractor} \\
\midrule
\# Context. & 0.85 & 0.79 & 0.92 & 0.89 \\
\# Influent. & 0.35 & 0.25 & 0.50 & 0.43 \\ 
\bottomrule
\end{tabular}
\caption{Average number of distractors containing contextual cues (\# Context.) and receiving a higher sum of CCI scores than all gold passages (\# Influent.), for Aya.
}
\label{tab:mirage_multi}
\end{table}

\subsection{Interpretability-based Evaluation}
\label{sec:mirage_multi}
We adopt once again MIRAGE \citep{qi-etal-2024-model} to understand how the internal model dynamics are affected by our various simulated multi-passage mRAG scenarios.
We sample 50 instances from each dataset and use MIRAGE to attribute Aya-expanse-8B responses to the provided passages via contrastive feature attribution \citep{yin-neubig-2022-interpreting}. 
Then, we compute 
\textbf{\# Contextual}: the average number of distracting passages that contain at least one contextual cue for the produced answer (i.e. a token marked by CCI in MIRAGE), and \textbf{\# Influential}: the average number of distractors that receive a higher sum of CCI attribution scores than all gold passages for each query.


The results in Table~\ref{tab:mirage_multi} support our observation that distractors exert a comparable effect regardless of their language, however in-language distractors have a slightly stronger effect.
When considering the sum of attribution scores given to the distractors compared with the gold passages, 
the difference becomes more noticeable (e.g., Aya tends to pay more attention to in-language distractors for MKQA when there is 1 distractor, compared to out-language ones). 


Taken together, our results indicate that the number of distractors can be more harmful for mRAG accuracy than the language in which those distractors are provided, when it comes to open-domain QA. 
On the multi-choice task, the negative effect of distractors is notably smaller and barely dependent on the passage language. 

\section{Conclusion}
In this work, we explored the challenge of consistent context utilization in mRAG systems. Specifically, we assessed the ability of various state-of-the-art LLMs to handle various kinds of multilingual context while strictly controlling for retrieval quality.
Our experiments across three diverse QA datasets, using standard accuracy evaluation as well as feature attribution analysis, reveal a remarkable ability of LLMs to understand multilingual contexts and to locate the important information in relevant passages regardless of their language.
In fact, models provided with multiple gold passages in languages different from that of the query are more likely to answer correctly 
than when provided with a single gold passage in the query language, reflecting the potential of retrieving cross-lingually rather than monolingually for mRAG. 

At the same time, we also detected some important directions for future improvement.
Firstly, poor generation abilities in many languages push the models to respond in a different language than that of the query, resulting in answers that would be deemed useless by most end-users. Importantly, we showed that this also happens when the retrieval works optimally. This suggests that, rather than just trying to optimize the retriever, it may be more effective to invest on the model generation abilities in a specific (set of) user language(s) --for instance by continued pre-training \citep{fujii2024continual, gao2024multilingual} on generic corpora of those languages-- or to apply techniques that push the model to decode in a given language, such as contrastive decoding \citep{li-etal-2023-contrastive, o2023contrastive}. 
Secondly, the presence of distracting passages (i.e., relevant to the query topic, but not directly functional to answer it) in the context can have a very negative effect on answer accuracy in open-domain QA. While this effect is rather similar regardless of the distractors' language, it does highlight the importance of carefully ranking the retrieved passages and to aim for precision when selecting which passages are provided to the model. 

To conclude, our work underscores the potential of cross-lingual retrieval in enhancing multilingual QA performance,
and stresses the importance of focusing not only on retrieval optimization but also on improving language-specific generation.
We believe this dual focus will be key to unlocking more robust and user-friendly mRAG systems that can operate effectively across diverse language settings. 



\section*{Limitations}

Our work relies on strict lexical matching to compute model answer accuracy and to detect gold passages. 
While commonly used, this approach is sensitive to minor variations or rephrasings of the answers and led to a serious underestimation of model performance with out-language gold passages in one of our QA datasets, XQUAD.
In our paper, we have tested and reported BERTScore and LLM-based evaluation on XQUAD, as detailed in Section \ref{sec:metrics} to enhance the robustness of our findings. These semantic evaluations mirror the trends observed with the lexical metric, mitigating—if not eliminating—the risk that paraphrasing may influence the results. Nevertheless, future work could incorporate broader metrics and benchmarks to make the assessment more comprehensive.

Additionally, the use of lexical matching in detecting gold passages may overlook passages that provide valuable information but in a slightly rephrased form compared to the gold answer.
Nonetheless, Table \ref{tab:multi_passage} shows that attaching even a single distracting passage identified by \textit{this} heuristic method substantially degrades model accuracy. Thus, despite its limitations, lexical matching proves to be a practical and effective way for locating distracting passages in our experimental setting. Future work could explore more semantic retrieval methods to capture paraphrased gold evidence.

On the retrieval side, simulating cross-language retrieval by combining results of N in-language retrievers may yield a more comprehensive set of passages than what we could obtain from a single run of a cross-language retriever.
While this does not affect our results on the side of context utilization, it may overestimate retriever performance when our findings are applied to real-world mRAG systems. 
In terms of datasets, XQUAD was the only one including parallel gold passages, which allowed us to fully isolate the effect of a passage language from that of its content. However, its extractive QA nature makes it less representative of realistic mRAG tasks, highlighting the need to develop better parallel mRAG datasets in future work.

\section*{Acknowledgments}

The authors have received funding from the Dutch Research Council (NWO): 
JQ is supported by NWA-ORC project LESSEN (grant nr. NWA.1389.20.183),
AB is supported by the above as well as NWO Talent Programme (VI.Vidi.221C.009),
RF is supported by the European Research Council (ERC) under European Union's Horizon 2020 programme (No.~819455).

\bibliography{anthology, custom}

\appendix

\section{Dataset Examples}
\label{sec:dataset_examples}
Examples of instances in each dataset are shown in Table \ref{tab:dataset_examples}.

\section{Full Statistics of the Filtered MKQA and GMMLU Datasets}
\label{sec:full_statistics}
The full statistics of the filtered MKQA and GMMLU datasets are shown in Table \ref{tab:dataset_statistic_details_full}.

\begin{table*}[!ht]
\centering
\small
\resizebox{0.99\linewidth}{!}{
\begin{tabular}{l|p{5.0cm}|p{5.0cm}|p{2.0cm}}
\toprule
\textbf{Dataset} & \textbf{Context provided in the dataset} & \textbf{Query} & \textbf{Gold Answer} \\
\midrule
XQUAD & The Panthers defense gave up just 308 points, ranking sixth in the league, while also leading the NFL in interceptions with 24 and boasting four Pro Bowl selections. ... also racking up 88 tackles and Pro Bowl cornerback Josh Norman, who developed into a shutdown corner during the season and had four interceptions, two of which were returned for touchdowns. & How many points did the Panthers defense surrender? & 308 \\
\midrule
MKQA & - & How long did it take the twin towers to be built? & 11.0 years \\
\midrule
GMMLU-Open & - & Which god supplanted the earlier Mesopotamian supreme god Enlil? & Marduk \\
\midrule
GMMLU-Choice & - & Which god supplanted the earlier Mesopotamian supreme god Enlil? A.Horus B.Inanna C.Marduk D.Isis & C \\
\bottomrule
\end{tabular}
}
\caption{Examples of instances in each dataset.}
\label{tab:dataset_examples}
\end{table*}

\begin{table*}[!ht]
\centering
\small
\resizebox{0.99\linewidth}{!}{
\begin{tabular}{l|cccccccccccccc}
\toprule
\rowcolor{black!10} \multicolumn{15}{c}{\bf MKQA (Total 5951 Questions = \# Inlang + \# Outlang - \# Both)} \\
\midrule
Query Lang. & en & it & es & de & fr & pt & nl & sv & ru & fi & ja & pl  \\
\midrule
\# Q. w/ Inlang & 5331 & 4466 & 4384 & 4352 & 4302 & 4133 & 4108 & 3984 & 3800 & 3639 & 3603 & 3594 \\
\# Q. w/ Outlang & 5787 & 5910 & 5942 & 5946 & 5944 & 5947 & 5947 & 5940 & 5945 & 5946 & 5944 & 5945 \\
\# Overlap & 5167 & 4425 & 4375 & 4347 & 4295 & 4129 & 4104 & 3973 & 3794 & 3634 & 3596 & 3588 \\
\midrule
Query Lang. & no & tr & hu & da & vi & he & ar & ms & ko & th & zh & km \\
\midrule
\# Q. w/ Inlang & 3515 & 3515 & 3482 & 3390 & 3365 & 3343 & 2986 & 2937 & 2934 & 2539 & 2537 & 703 \\
\# Q. w/ Outlang & 5949 & 5945 & 5943 & 5946 & 5951 & 5946 & 5948 & 5942 & 5947 & 5945 & 5948 & 5950 \\
\# Overlap & 3513 & 3509 & 3474 & 3385 & 3365 & 3338 & 2983 & 2928 & 2930 & 2533 & 2534 & 702 \\
\midrule
\rowcolor{black!10} \multicolumn{15}{c}{\bf GMMLU (Total 4136 Questions = \# Inlang + \# Outlang - \# Both)} \\
\midrule
Query Lang. & en & ja & it & id & ko & nl & zh & vi & sv & pt & de & tr & ro & cs \\
\midrule
\# Q. w/ Inlang & 2588 & 2054 & 1864 & 1778 & 1725 & 1712 & 1695 & 1689 & 1688 & 1679 & 1611 & 1583 & 1513 & 1512 \\
\# Q. w/ Outlang & 4040 & 4064 & 4118 & 4115 & 4097 & 4125 & 4094 & 4116 & 4124 & 4118 & 4111 & 4121 & 4126 & 4116 \\
\# Overlap & 2492 & 1982 & 1846 & 1757 & 1686 & 1701 & 1653 & 1669 & 1676 & 1661 & 1586 & 1568 & 1503 & 1492 \\
\midrule
Query Lang. & ru & es & ms & pl & uk & fr & ar & fa & el & sr & he & hi & fil & lt \\
\midrule
\# Q. w/ Inlang & 1503 & 1502 & 1464 & 1462 & 1422 & 1415 & 1373 & 1350 & 1317 & 1288 & 1160 & 1142 & 1125 & 1071 \\
\# Q. w/ Outlang & 4126 & 4109 & 4126 & 4124 & 4130 & 4122 & 4118 & 4125 & 4130 & 4130 & 4118 & 4133 & 4130 & 4132 \\
\# Overlap & 1493 & 1475 & 1454 & 1450 & 1416 & 1401 & 1355 & 1339 & 1311 & 1282 & 1142 & 1139 & 1119 & 1067 \\
\midrule
Query Lang. & bn & ky & ha & te & sw & ig & si & ne & am & ny & mg & so & sn & yo \\
\midrule
\# Q. w/ Inlang & 1005 & 985 & 930 & 924 & 923 & 831 & 792 & 746 & 650 & 634 & 625 & 559 & 497 & 389 \\
\# Q. w/ Outlang & 4125 & 4121 & 4123 & 4130 & 4129 & 4125 & 4132 & 4132 & 4135 & 4129 & 4133 & 4129 & 4134 & 4129 \\
\# Overlap & 994 & 970 & 917 & 918 & 916 & 820 & 788 & 742 & 649 & 627 & 622 & 552 & 495 & 382 \\
\bottomrule
\end{tabular}
}
\caption{The statistics of the filtered subset of MKQA and Global-MMLU where each query has gold passages in at least one studied language. For all languages, there is a portion of queries where useful information can only be found in out-language passages, which is particularly evident in low-resource languages. \# Inlang: Number of queries having gold passages retrieved from the corpora of the query language. \# Outlang: Number of queries having out-language gold passages. I.e. useful information is stored in the corpora of languages other than the query language. \# Overlap: Number of queries that have useful information retrieved from both in-language and out-language corpora.
}
\label{tab:dataset_statistic_details_full}
\end{table*}

\section{Subjects Covered by the Filtered GMMLU Set}
\label{sec:covered_subjects}
As shown in Table \ref{tab:GMMLU_category}, 55 subjects belonging to 6 categories are covered by the filtered set of Global-MMLU, which ensures the diversity of the instances evaluated in our experiments.

\begin{table*}[!ht]
\centering
\small
\resizebox{0.99\linewidth}{!}{
\begin{tabular}{c|p{12.0cm}}
\toprule
\textbf{Category} & \textbf{Subject} \\
\midrule
STEM & high\_school\_computer\_science, high\_school\_statistics, computer\_security, college\_biology, college\_chemistry, machine\_learning, high\_school\_mathematics, elementary\_mathematics, college\_mathematics, electrical\_engineering, college\_physics, astronomy, conceptual\_physics, high\_school\_chemistry, high\_school\_physics, high\_school\_biology, college\_computer\_science, anatomy \\
\midrule
Business & business\_ethics, management, marketing, professional\_accounting \\
\midrule
Medical & professional\_medicine, virology, college\_medicine, clinical\_knowledge, human\_aging, medical\_genetics, nutrition \\
\midrule
Social Sciences & high\_school\_psychology, econometrics, sociology, high\_school\_microeconomics, high\_school\_geography, public\_relations, security\_studies, professional\_psychology, high\_school\_government\_and\_politics, high\_school\_macroeconomics, human\_sexuality, us\_foreign\_policy \\
\midrule
Humanities & international\_law, high\_school\_world\_history, moral\_disputes, prehistory, world\_religions, jurisprudence, high\_school\_us\_history, philosophy, professional\_law, formal\_logic, logical\_fallacies, high\_school\_european\_history \\
\midrule
Other & miscellaneous, global\_facts \\
\bottomrule
\end{tabular}
}
\caption{The categories and subjects covered by the filtered GMMLU.}
\label{tab:GMMLU_category}
\end{table*}

\section{Extensive Evaluation}
\label{app:extensive_eval}

\begin{table}[!t]
\small
\centering
\begin{tabular}{c|cc|cc}
\toprule
\multirow{2}{*}{\bf Lang.} & \multicolumn{2}{c|}{\bf BERTScore} & \multicolumn{2}{c}{\bf LLM-Based Score}\\
& \bf IN. & \bf OUT. & \bf IN. & \bf OUT. \\
\midrule
en     & 90.27 & 82.58 & 93.19 & 81.54 (+12.93) \\
ar      & 82.87 & 81.55 & 91.93 & 63.98 (+23.25) \\
de      & 81.60 & 80.69 & 90.25 & 71.73 (+14.86) \\
el       & 82.38 & 81.24 & 92.18 & 66.00 (+21.04) \\
es     & 82.15 & 81.14 & 94.37 & 73.05 (+14.62) \\
hi       & 83.31 & 82.15 & 89.75 & 65.78 (+16.71) \\
ro    & 81.68 & 80.65 & 91.93 & 69.11 (+18.69) \\
ru     & 83.22 & 81.95 & 91.51 & 66.78 (+21.18) \\
th        & 84.21 & 83.23 & 86.81 & 57.78 (+23.02) \\
tr     & 81.31 & 80.18 & 88.40 & 63.39 (+23.37) \\
vi  & 82.81 & 81.62 & 89.75 & 65.81 (+23.05) \\
zh     & 84.07 & 82.95 & 90.34 & 65.39 (+21.46) \\
\bottomrule
\end{tabular}
\caption{BERTScore (F1) and LLM-based evaluation (Accuracy) on XQUAD with AYA. The numbers between brackets indicate the proportion of queries that are correctly answered but in the wrong language.}
\label{tab:extensive_eval}
\end{table}

Since exact matching could be overly strict for the evaluation, we further adopt two complementary metrics on XQUAD with AYA. 

\paragraph{Semantic similarity (BERTScore)} We compute BERTScore, serving as a language-agnostic metric, between each model response and its ground-truth answer based on the semantic similarity of model responses with the gold answer. Table \ref{tab:extensive_eval} shows that models achieve comparable F1 scores in all query languages when fed gold passages in- or out-language. This finding is in line with our claim that LLMs are capable of understanding the gold passages regardless of their languages. 

\paragraph{LLM-based evaluation (GPT-4.1-nano)}
However, semantic similarity cannot capture language mismatching. Therefore, we prompted GPT-4.1-nano to judge whether each response matches (i) the correct answer and (ii) its translation in the passage language. As shown in Table \ref{tab:extensive_eval}, overall accuracy on board is higher than lexical-matching accuracy in our paper, but the trend remains: models score better on IN than on OUT. If we allow “correct answer in the wrong language” as acceptable, the IN/OUT gap almost disappears.

Taken together, both semantic and LLM-based evaluation support our claim that LLMs are able to understand the multilingual gold passages regardless of their languages, but suffer from decoding the answer correctly in the user query language.

\section{Prompts and Instructions}
\label{sec:prompts}
To ensure the model responses are always in the query language, we follow previous works \citep{chirkova-etal-2024-retrieval, zhang-etal-2024-respond} and adopt language-specific instructions to explicitly and implicitly guide the model to generate responses in the user-readable language. The examples in English, Spanish, and Chinese are listed in Table \ref{tab:instructions_open} and Table \ref{tab:instructions_choice}.

\begin{CJK*}{UTF8}{gbsn}
\begin{table*}[!ht]
\centering
\small
\resizebox{0.99\linewidth}{!}{
\begin{tabular}{c|c|p{12.0cm}}
\toprule
\textbf{Language} & \textbf{Setup} & \textbf{Instruction} \\
\midrule
en & No Ctx & Write a high-quality answer to the given question. Please respond in English. \\
\cmidrule{2-3}
& Ctx & Write a high-quality answer to the given question using the provided search results. Please respond in English. \\
\midrule
es & No Ctx & Escriba una respuesta de alta calidad a la pregunta planteada. Por favor responda en español. \\
\cmidrule{2-3}
& Ctx & Escriba una respuesta de alta calidad a la pregunta planteada utilizando los resultados de búsqueda proporcionados. Por favor, responda en español. \\
\midrule
zh & No Ctx & 请对所给问题写出高质量的答案。请使用中文回答。 \\
\cmidrule{2-3}
& Ctx & 使用提供的搜索结果对给定的问题写出高质量的答案。请用中文回答。 \\
\bottomrule
\end{tabular}
}
\caption{The examples of the adopted instructions for guiding LLMs to generate responses in the user languages on the open QA tasks (XQUAD, MKQA, GMMLU-Open).}
\label{tab:instructions_open}
\end{table*}
\end{CJK*}

\begin{CJK*}{UTF8}{gbsn}
\begin{table*}[!ht]
\centering
\small
\resizebox{0.99\linewidth}{!}{
\begin{tabular}{c|p{12.0cm}}
\toprule
\textbf{Language} & \textbf{Instruction} \\
\midrule
en & Please choose the most suitable one among A, B, C and D as the answer to the question, and return it in the following format: \\ 
& [choice]\\
& where [choice] must be one of [A], [B], [C] and [D]. \\
\midrule
es & Elija la respuesta más adecuada entre A, B, C y D a la pregunta y devuélvala en el siguiente formato: \\
& [opción] \\
& donde [opción] debe ser una de [A], [B], [C] y [D]. \\
\midrule
zh & 请在 A、B、C 和 D 中选择最合适的一个作为问题的答案，并按照以下格式返回：\\
& [choice] \\
& 其中 [choice] 必须是 [A]、[B]、[C] 和 [D] 之一。 \\
\bottomrule
\end{tabular}
}
\caption{The examples of the adopted instructions for guiding LLMs to generate responses in the user languages on the multi-choice QA task (GMMLU-Choice).}
\label{tab:instructions_choice}
\end{table*}
\end{CJK*}

\section{Error Analysis on XQUAD}
\label{sec:error_analysis}
\begin{CJK*}{UTF8}{gbsn}
While our MKQA and GMMLU results strongly suggest our studied LLMs can understand the provided passages regardless of their language, the in-/out-language gap in XQUAD remains unexplained.
To address this, we conduct a manual error analysis on XQUAD with Aya-Expanse-8B, focusing on a random sample of 20 Spanish and 20 Chinese queries that were answered correctly when provided with in-language passages, but wrongly with out-language passages. In most cases, we observe that models successfully understood the context and generated a proper response, however, this response did not perfectly match the gold answer provided in the dataset.
This can be due to the presence of synonyms or slight paraphrases of the gold answer, or --in the case of languages with different scripts-- to transliteration variations \citep{knight-graehl-1997-machine}.
For instance, the gold answer for a Spanish question is \textit{`evolución de la lengua y la literatura alemanas'} (i.e. `evolution of the German language and literature'). In the in-language setup, the model manages to generate this exact string as it is included in the provided Spanish passage.
However, when the same passage is provided in English, the model generates the semantically equivalent phrase \textit{`... evolución del idioma y la literatura alemana...'}, or \textit{`...desarrollo del idioma y la literatura alemana...'} when the passage is provided in Chinese.
Similarly, for a Chinese query with gold answer \textit{`亚里士多德宇宙学'} (i.e. `Aristotelian cosmology'), model responses slightly differ when provided with different out-language passages (e.g. `亚里士多德宇宙论', `阿里斯托的宇宙论', or `阿里斯托特利宇宙论' with English, Arabic, or Greek passage respectively), all of which are correct translations of `Aristotelian cosmology'.
\end{CJK*}
While this issue can always affect lexical-matching evaluation, it is particularly severe in XQUAD as many answers in this dataset are named entities or sentence segments due to the extractive nature of the task, which in turn causes an underestimation of the models capability.

\section{Advanced System Prompting}
\label{app:advanced_prompt}
In our main experiments, we follow the previous works \citep{chirkova-etal-2024-retrieval, zhang-etal-2024-respond} and adopt the direct prompt. To test if a stronger prompt could mitigate language-mismatch errors, we add a two-step instruction that first allows the model to answer in any appropriate language, then explicitly translates the answer into the query language. Formally: `Write a high-quality answer to the given question using the provided search results. Please respond in English. Specifically, please follow the two steps below. Step 1: Generate a complete answer to the question in any appropriate language. Step 2: Translate your entire answer into clear, natural‐sounding English.'

Same as the main experiment in the paper, the prompt is translated into other query languages and explicitly specifies the desired generation language. For instance, the prompt for Spanish queries is: 

`Escriba una respuesta de alta calidad a la pregunta dada utilizando los resultados de búsqueda proporcionados. Por favor responda en español. Específicamente, siga los dos pasos a continuación. Paso 1: Genere una respuesta completa a la pregunta en cualquier idioma apropiado. Paso 2: Traduce toda tu respuesta a un español claro y con sonido natural.'

\begin{table}[!t]
\centering
\begin{tabular}{c|cc}
\toprule
\bf \multirow{2}{*}{Language} & \multicolumn{2}{c}{\bf LLM-Based Score}\\
& \bf IN. & \bf OUT. \\
\midrule
en     & 94.71 & 82.52 (+11.24) \\
ar      & 87.14 & 64.38 (+24.85) \\
de      & 89.92 & 73.12 (+13.76) \\
el       & 90.59 & 65.11 (+22.05) \\
es     & 92.61 & 73.51 (+14.23) \\
hi       & 85.13 & 63.80 (+20.06) \\
ro    & 92.61 & 67.00 (+19.17) \\
ru     & 90.67 & 69.49 (+19.34) \\
th        & 86.89 & 60.04 (+25.12) \\
tr     & 85.46 & 63.46 (+21.35) \\
vi  & 91.34 & 67.16 (+22.85) \\
zh     & 90.84 & 69.60 (+20.28) \\
\bottomrule
\end{tabular}
\caption{LLM-based evaluation score (Accuracy) on XQUAD with Aya, where stronger multi-step reasoning prompts are adopted. Nonetheless, the language-mismatching issue persists.}
\label{tab:stronger_prompt}
\end{table}


We run this prompt on XQUAD with AYA and evaluate via GPT-4.1-nano, the same setups and LLM-based evaluation as above in Appendix \ref{app:extensive_eval}. As shown in Table \ref{tab:stronger_prompt}, compared to the original prompts, these stronger instructions reduced, but did not eliminate, the gap between in-language and out-language accuracy. Specifically, many responses still contained correct answers but remained in the wrong passage language, indicating that even explicitly guiding the LLM to do `think then translation' cannot fully resolve decoding failures. These results underscore that decoding, rather than understanding, remains a substantial bottleneck.

\begin{table*}[!t]
\centering
\begin{tabular}{c|rrlll}
\toprule
\multirow{2}{*}{\bf Language} & \multicolumn{5}{c}{\bf Accuracy (Gemma3-27B-IT)} \\
\cmidrule{2-6}
& \textbf{Non.} & \textbf{In.} & \textbf{Out. (AVG)} & \textbf{Out. (Best)} & \textbf{Out. (Worst)} \\
\midrule
en & 21.3 & 86.7 & 67.1 (+4.9) & 72.9 (+5.5) & 60.3 (+9.4)  \\
es & 16.6 & 72.9 & 53.2 (+5.8) & 59.2 (+9.4) & 48.2 (+7.6)  \\
de & 17.0 & 72.5 & 52.0 (+4.7) & 56.9 (+7.2) & 48.2 (+7.0)  \\
ro & 14.5 & 76.2 & 52.0 (+4.3) & 56.6 (+8.3) & 47.4 (+6.2)  \\
vi & 15.1 & 77.6 & 49.9 (+7.4) & 53.7 (+13.4) & 44.3 (+9.7)  \\
tr & 12.2 & 67.4 & 45.4 (+5.9) & 50.5 (+9.8) & 42.1 (+5.0)  \\
el & 10.4 & 68.9 & 40.9 (+8.3) & 44.7 (+16.1) & 38.2 (+6.8)  \\
zh & 13.9 & 79.2 & 44.1 (+13.7) & 45.7 (+20.7) & 41.5 (+7.3)  \\
ar & 8.8 & 65.8 & 35.5 (+9.5) & 37.4 (+19.6) & 32.3 (+5.8)  \\
hi & 13.3 & 74.8 & 42.6 (+6.0) & 46.8 (+15.0) & 38.7 (+4.2)  \\
ru & 11.4 & 66.6 & 40.6 (+9.7) & 42.6 (+11.8) & 36.9 (+6.7)  \\
th & 11.1 & 74.6 & 35.4 (+14.2) & 36.4 (+30.0) & 34.5 (+9.8)  \\
\midrule
\multirow{2}{*}{\bf Language} & \multicolumn{5}{c}{\bf Accuracy (GPT5-nano)} \\
\cmidrule{2-6}
& \textbf{Non.} & \textbf{In.} & \textbf{Out. (AVG)} & \textbf{Out. (Best)} & \textbf{Out. (Worst)} \\
\midrule
en & 25.6 & 74.0 & 55.1 (+7.1) & 58.3 (+8.3) & 51.8 (+7.8)  \\
es & 20.2 & 64.8 & 50.2 (+3.6) & 55.1 (+7.2) & 46.3 (+2.4)  \\
de & 20.1 & 60.2 & 48.0 (+1.9) & 52.0 (+6.2) & 44.9 (+0.8)  \\
ro & 19.2 & 59.8 & 46.1 (+2.6) & 49.4 (+7.3) & 42.9 (+1.4)  \\
vi & 18.3 & 58.8 & 45.1 (+4.4) & 47.8 (+5.0) & 42.4 (+2.1)  \\
tr & 15.5 & 49.2 & 38.0 (+5.6) & 39.8 (+7.4) & 35.3 (+2.4)  \\
el & 12.5 & 55.5 & 35.6 (+6.1) & 39.3 (+16.2) & 32.9 (+2.1)  \\
zh & 16.5 & 59.6 & 39.1 (+8.9) & 41.4 (+18.2) & 37.8 (+3.6)  \\
ar & 10.3 & 48.2 & 31.3 (+8.8) & 33.1 (+7.1) & 30.0 (+3.9)  \\
hi & 12.8 & 45.2 & 29.8 (+10.3) & 32.3 (+22.6) & 27.0 (+13.3)  \\
ru & 12.8 & 48.9 & 34.1 (+6.4) & 36.4 (+14.0) & 31.8 (+3.9)  \\
th & 11.9 & 53.4 & 32.6 (+10.6) & 34.2 (+25.2) & 30.4 (+2.1)  \\
\bottomrule
\end{tabular}
\caption{Language-specific results on XQUAD with larger LLMs (Gemma3-27B-IT and GPT5-nano). Numbers between brackets indicate the proportion of queries that are correctly answered but in the wrong language (i.e., not the query language).}
\label{tab:larger_models}
\end{table*}

\section{Extended Evaluation on Larger Models}
\label{app:larger_model}
To enhance the robustness of our experiments, we repeat the XQuAD evaluation (using the same setup) on a 27B open-source model (Gemma3-27B-IT) and a closed-source model estimated at 8-18B parameters (GPT5-nano)\footnote{\url{https://www.r-bloggers.com/2025/08/how-many-parameters-does-gpt-5-have/}}. The results in Table \ref{tab:larger_models} show that, although overall accuracy of out-language passages improves, it remains substantially lower than on in-language passages. Moreover, a non-negligible fraction of questions are answered correctly in content but produced in the wrong language when the model receives out-language passages. These findings are consistent with Section \ref{sec:accuracy_results} and further strengthen the generalization of our findings.

\section{Language-specific Results}
\label{sec:full_results}
\begin{figure*}[!t]
    \centering
    \includegraphics[width=0.99\textwidth]{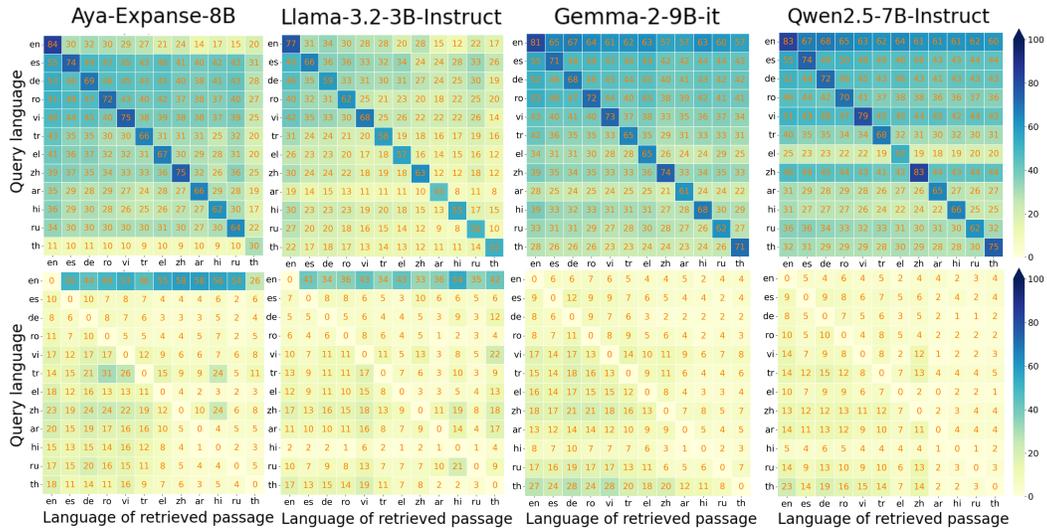}
    \caption{\small Model performance on XQUAD when the query is concatenated with passage in each studied language. Top: The portion of queries that can be correctly answered in the user language. Bottom: The portion of queries for which the LLMs generate the correct answer but in the wrong (passage) language. For a part of correctly answered queries, the gold answers are the same words in the passage and query languages. In these cases, we only consider them in the above heatmaps to ensure that there is no overlapping between the two vertical heatmaps and that they are addable.}
    \label{fig:XQUAD_heatmap}
\end{figure*}

The detailed results for each query-passage language pair on XQUAD are given in Figure \ref{fig:XQUAD_heatmap}. The detailed single-passage mRAG results for each language on all datasets are provided in Table \ref{tab:full_XQUAD_1} to \ref{tab:full_GMMLU_choice}. The detailed results for each language in the multi-passage mRAG experiments are shown in Table \ref{tab:multi_mirage_full_1} to \ref{tab:multi_mirage_full_4}.

\begin{table*}[!t]
\centering
\begin{tabular}{c|rrlll}
\toprule
\multirow{2}{*}{\bf Language} & \multicolumn{5}{c}{\bf Accuracy (Aya)} \\
\cmidrule{2-6}
& \textbf{Non.} & \textbf{In.} & \textbf{Out. (AVG)} & \textbf{Out. (Best)} & \textbf{Out. (Worst)} \\
\midrule
en & 19.4 & 83.7 & 23.6 (+49.9) & 32.4 (+44.1) & 14.0 (+58.5)  \\
es & 13.9 & 73.5 & 43.7 (+6.3) & 54.9 (+10.1) & 30.9 (+5.5)  \\
de & 12.2 & 69.2 & 42.5 (+5.0) & 53.4 (+7.6) & 28.1 (+4.4)  \\
ro & 12.0 & 72.4 & 40.9 (+5.8) & 51.1 (+10.8) & 26.8 (+5.3)  \\
vi & 14.3 & 75.0 & 39.4 (+10.6) & 48.2 (+16.8) & 25.5 (+8.4)  \\
tr & 9.1 & 66.1 & 31.2 (+16.4) & 43.0 (+13.9) & 19.9 (+10.8)  \\
el & 8.5 & 67.5 & 31.5 (+8.6) & 40.5 (+17.8) & 20.0 (+2.6)  \\
zh & 12.6 & 74.6 & 33.3 (+17.3) & 39.0 (+22.6) & 25.5 (+8.2)  \\
ar & 7.6 & 66.3 & 27.4 (+12.1) & 35.0 (+19.8) & 18.7 (+5.3)  \\
hi & 6.1 & 62.4 & 27.3 (+9.4) & 36.0 (+15.5) & 17.2 (+2.6)  \\
ru & 9.2 & 63.8 & 29.6 (+11.0) & 34.1 (+16.7) & 22.1 (+5.2)  \\
th & 2.1 & 30.3 & 10.0 (+9.3) & 11.4 (+17.6) & 9.1 (+5.8)  \\
\midrule
AVG & 10.6 & 67.1 & 31.7 (+13.5) & 39.9 (+17.8) & 21.5 (+10.2)\\
\midrule
\multirow{2}{*}{\bf Language} & \multicolumn{5}{c}{\bf Accuracy (Llama)} \\
\cmidrule{2-6}
& \textbf{Non.} & \textbf{In.} & \textbf{Out. (AVG)} & \textbf{Out. (Best)} & \textbf{Out. (Worst)} \\
\midrule
en & 14.4 & 76.8 & 24.2 (+39.1) & 33.7 (+34.1) & 12.1 (+53.6)  \\
es & 8.8 & 66.2 & 32.3 (+6.5) & 49.2 (+7.1) & 24.0 (+6.0)  \\
de & 7.6 & 59.0 & 30.1 (+5.5) & 45.7 (+5.0) & 19.0 (+12.0)  \\
ro & 6.6 & 61.8 & 25.2 (+4.4) & 40.2 (+5.8) & 18.2 (+1.4)  \\
vi & 8.5 & 68.4 & 27.1 (+9.6) & 41.8 (+9.7) & 14.4 (+21.9)  \\
tr & 5.1 & 57.9 & 20.5 (+9.2) & 30.7 (+13.1) & 15.8 (+8.6)  \\
el & 2.5 & 57.5 & 18.2 (+8.4) & 25.8 (+12.4) & 11.8 (+12.5)  \\
zh & 5.2 & 62.9 & 18.9 (+14.2) & 30.3 (+16.6) & 11.8 (+18.0)  \\
ar & 2.0 & 45.3 & 12.0 (+10.7) & 18.7 (+10.5) & 8.2 (+17.5)  \\
hi & 2.7 & 55.3 & 19.6 (+2.3) & 29.6 (+2.4) & 13.4 (+0.8)  \\
ru & 4.1 & 49.8 & 15.3 (+9.8) & 26.8 (+10.0) & 7.6 (+20.7)  \\
th & 2.9 & 51.8 & 14.6 (+10.1) & 21.6 (+16.9) & 11.0 (+1.8)  \\
\midrule
AVG & 5.9 & 59.4 & 21.5 (+10.8) & 32.8 (+12.0) & 13.9 (+14.6) \\
\bottomrule
\end{tabular}
\caption{Language-specific results on XQUAD with single-passage mRAG setup. Numbers between brackets indicate the proportion of queries that are correctly answered but in the wrong language (i.e., not the query language).}
\label{tab:full_XQUAD_1}
\end{table*}

\begin{table*}[!t]
\centering
\begin{tabular}{c|rrlll}
\toprule
\multirow{2}{*}{\bf Language} & \multicolumn{5}{c}{\bf Accuracy (Gemma)} \\
\cmidrule{2-6}
& \textbf{Non.} & \textbf{In.} & \textbf{Out. (AVG)} & \textbf{Out. (Best)} & \textbf{Out. (Worst)} \\
\midrule
en & 17.9 & 80.6 & 61.5 (+5.0) & 67.1 (+6.5) & 56.6 (+5.3)  \\
es & 12.4 & 70.8 & 46.3 (+6.6) & 54.5 (+9.4) & 41.7 (+4.1)  \\
de & 12.1 & 67.7 & 44.1 (+4.4) & 52.0 (+7.6) & 39.8 (+2.3)  \\
ro & 12.1 & 72.1 & 43.3 (+6.2) & 52.9 (+8.5) & 37.8 (+5.9)  \\
vi & 10.8 & 73.4 & 37.7 (+11.3) & 43.0 (+16.8) & 33.1 (+10.8)  \\
tr & 8.4 & 65.2 & 33.7 (+10.5) & 41.8 (+15.0) & 29.1 (+11.1)  \\
el & 5.0 & 65.0 & 28.2 (+10.1) & 34.0 (+15.7) & 23.7 (+4.0)  \\
zh & 9.3 & 73.8 & 34.5 (+13.7) & 38.9 (+17.6) & 32.9 (+6.9)  \\
ar & 4.9 & 61.3 & 24.3 (+9.8) & 27.6 (+12.9) & 20.7 (+9.0)  \\
hi & 6.1 & 67.7 & 31.2 (+5.5) & 38.9 (+7.6) & 26.7 (+3.4)  \\
ru & 7.4 & 62.1 & 28.9 (+12.1) & 32.9 (+16.8) & 26.1 (+5.5)  \\
th & 5.4 & 71.3 & 24.9 (+20.0) & 28.2 (+26.7) & 22.8 (+11.6)  \\
\midrule
AVG & 9.3 & 69.2 & 36.5 (+9.6) & 42.6 (+13.4) & 32.6 (+6.7)\\
\midrule
\multirow{2}{*}{\bf Language} & \multicolumn{5}{c}{\bf Accuracy (Qwen)} \\
\cmidrule{2-6}
& \textbf{Non.} & \textbf{In.} & \textbf{Out. (AVG)} & \textbf{Out. (Best)} & \textbf{Out. (Worst)} \\
\midrule
en & 23.5 & 82.7 & 63.1 (+3.9) & 67.6 (+4.4) & 60.3 (+3.7)  \\
es & 17.2 & 74.3 & 46.6 (+5.7) & 55.2 (+9.3) & 42.5 (+2.4)  \\
de & 15.2 & 72.0 & 44.9 (+4.1) & 51.2 (+8.3) & 41.4 (+1.7)  \\
ro & 13.7 & 70.1 & 39.1 (+4.9) & 46.2 (+9.8) & 35.5 (+1.9)  \\
vi & 15.7 & 79.1 & 45.7 (+6.2) & 50.5 (+14.2) & 42.1 (+2.4)  \\
tr & 9.8 & 68.4 & 32.9 (+8.9) & 40.1 (+13.5) & 29.8 (+3.5)  \\
el & 4.8 & 51.6 & 20.9 (+5.0) & 25.1 (+9.7) & 18.1 (+0.4)  \\
zh & 18.9 & 83.0 & 43.9 (+8.5) & 47.7 (+12.9) & 41.2 (+12.4)  \\
ar & 9.1 & 65.3 & 27.7 (+9.1) & 31.0 (+14.5) & 25.7 (+10.3)  \\
hi & 5.5 & 66.1 & 25.6 (+3.7) & 30.8 (+5.5) & 22.1 (+0.6)  \\
ru & 9.1 & 62.2 & 32.4 (+8.5) & 36.1 (+9.0) & 29.6 (+2.9)  \\
th & 8.5 & 75.1 & 29.5 (+11.7) & 32.2 (+22.6) & 27.6 (+14.4)  \\
\midrule
AVG & 12.6 & 70.8 & 37.7 (+6.7) & 42.8 (+11.1) & 34.7 (+4.7)\\
\bottomrule
\end{tabular}
\caption{Extension: Language-specific results on XQUAD with single-passage mRAG setup. Numbers between brackets indicate the proportion of queries that are correctly answered but in the wrong language (i.e., not the query language).}
\label{tab:full_XQUAD_2}
\end{table*}

\begin{table*}[!t]
\centering
\small
\begin{tabular}{c|rrl|rrl}
\toprule
\multirow{2}{*}{\bf Language} & \multicolumn{3}{c|}{\bf Accuracy (Aya)} & \multicolumn{3}{c}{\bf Accuracy (Llama)} \\
& \textbf{Non.} & \textbf{In.} & \textbf{Out.} & \textbf{Non.} & \textbf{In.} & \textbf{Out.} \\
\midrule
en  & 58.0 & 87.5 & 69.4 (+15.3)  & 61.8 & 82.4 & 63.9 (+10.2) \\
it  & 53.0 & 86.9 & 78.5 (+4.6)  & 44.7 & 70.6 & 63.6 (+3.7) \\
es  & 54.6 & 85.7 & 77.9 (+4.2)  & 46.9 & 72.1 & 66.8 (+3.1) \\
de  & 52.8 & 84.4 & 78.2 (+3.2)  & 46.0 & 70.6 & 65.3 (+2.2) \\
fr  & 55.4 & 86.9 & 79.1 (+3.1)  & 48.8 & 74.2 & 67.3 (+2.6) \\
pt  & 54.4 & 84.7 & 76.2 (+5.4)  & 45.0 & 70.1 & 63.8 (+3.9) \\
nl  & 55.9 & 85.6 & 76.8 (+4.2)  & 50.3 & 70.3 & 64.3 (+2.8) \\
sv  & 39.1 & 76.7 & 66.7 (+5.3)  & 44.9 & 69.7 & 63.3 (+3.4) \\
ru  & 43.6 & 79.6 & 59.7 (+14.0)  & 23.7 & 56.8 & 37.7 (+10.1) \\
fi  & 17.0 & 72.4 & 58.3 (+7.4)  & 27.2 & 62.4 & 53.9 (+4.2) \\
ja  & 46.8 & 82.1 & 54.9 (+23.3)  & 21.7 & 57.5 & 36.1 (+12.3) \\
pl  & 50.1 & 79.2 & 68.0 (+6.1)  & 35.3 & 60.6 & 51.5 (+4.4) \\
no  & 39.9 & 76.4 & 64.1 (+7.9)  & 43.6 & 66.4 & 58.3 (+5.3) \\
tr  & 52.0 & 82.3 & 72.2 (+7.6)  & 38.2 & 66.3 & 55.5 (+5.5) \\
hu  & 26.7 & 67.5 & 52.6 (+8.3)  & 34.4 & 59.3 & 49.2 (+5.1) \\
da  & 42.9 & 75.8 & 66.2 (+6.1)  & 46.4 & 66.5 & 60.3 (+4.2) \\
vi  & 54.4 & 79.0 & 71.2 (+6.5)  & 47.0 & 67.3 & 60.2 (+5.1) \\
he  & 36.2 & 78.5 & 46.5 (+25.4)  & 5.9 & 25.6 & 17.1 (+20.3) \\
ar  & 38.8 & 72.5 & 50.4 (+19.5)  & 17.2 & 46.4 & 27.2 (+10.7) \\
ms  & 53.6 & 78.4 & 68.4 (+7.7)  & 46.8 & 62.5 & 55.5 (+5.0) \\
ko  & 15.3 & 30.7 & 25.8 (+25.3)  & 19.2 & 52.9 & 30.2 (+17.6) \\
th  & 15.0 & 40.5 & 25.8 (+16.7)  & 24.2 & 46.9 & 32.0 (+13.3) \\
zh  & 48.4 & 78.0 & 56.6 (+20.2)  & 27.8 & 52.4 & 37.4 (+16.3) \\
km  & 7.8 & 19.2 & 14.9 (+8.2)  & 6.0 & 13.0 & 9.4 (+0.2) \\
\midrule
AVG & 42.2 & 73.8 & 60.8 (+10.6) & 35.5 & 60.1 & 49.6 (+7.1) \\
\midrule
\multirow{2}{*}{\bf Language} & \multicolumn{3}{c|}{\bf Accuracy (Gemma)} & \multicolumn{3}{c}{\bf Accuracy (Qwen)} \\
& \textbf{Non.} & \textbf{In.} & \textbf{Out.} & \textbf{Non.} & \textbf{In.} & \textbf{Out.} \\
\midrule
en  & 65.0 & 79.8 & 75.7 (+1.3)  & 55.6 & 86.0 & 81.7 (+1.1) \\
it  & 59.6 & 80.3 & 75.4 (+3.7)  & 44.5 & 81.2 & 72.7 (+4.1) \\
es  & 61.4 & 78.0 & 73.3 (+3.8)  & 48.4 & 80.9 & 73.5 (+3.6) \\
de  & 58.3 & 75.9 & 73.7 (+2.4)  & 44.6 & 78.2 & 70.5 (+2.9) \\
fr  & 61.8 & 80.7 & 74.8 (+2.6)  & 49.0 & 82.3 & 74.9 (+2.6) \\
pt  & 60.1 & 76.5 & 70.5 (+5.0)  & 48.5 & 78.6 & 71.4 (+4.2) \\
nl  & 64.5 & 74.8 & 71.1 (+3.3)  & 45.7 & 79.0 & 71.1 (+3.3) \\
sv  & 59.9 & 73.5 & 70.2 (+3.6)  & 43.7 & 75.4 & 68.0 (+4.6) \\
ru  & 42.8 & 68.8 & 50.6 (+10.6)  & 30.7 & 71.0 & 51.2 (+10.3) \\
fi  & 44.6 & 68.3 & 62.4 (+4.6)  & 23.9 & 72.5 & 60.9 (+5.7) \\
ja  & 42.7 & 69.9 & 47.9 (+13.7)  & 31.1 & 75.2 & 46.3 (+17.5) \\
pl  & 55.0 & 67.8 & 62.0 (+4.6)  & 36.9 & 68.9 & 59.1 (+5.4) \\
no  & 59.3 & 68.6 & 63.9 (+5.8)  & 42.3 & 73.8 & 63.8 (+6.9) \\
tr  & 56.7 & 66.8 & 62.8 (+4.7)  & 36.4 & 72.4 & 64.4 (+5.7) \\
hu  & 51.2 & 64.9 & 61.9 (+5.5)  & 27.4 & 67.1 & 55.9 (+6.9) \\
da  & 61.1 & 66.8 & 64.4 (+4.9)  & 44.8 & 72.7 & 65.8 (+5.1) \\
vi  & 54.4 & 66.5 & 64.1 (+5.3)  & 50.9 & 74.3 & 69.3 (+4.9) \\
he  & 29.7 & 65.3 & 39.5 (+16.6)  & 19.9 & 66.6 & 34.3 (+18.2) \\
ar  & 30.0 & 61.1 & 42.8 (+9.9)  & 27.9 & 64.8 & 42.3 (+12.7) \\
ms  & 60.5 & 63.8 & 62.3 (+6.9)  & 47.1 & 68.1 & 63.4 (+6.4) \\
ko  & 12.2 & 17.5 & 19.7 (+9.7)  & 23.3 & 59.0 & 39.1 (+16.6) \\
th  & 40.0 & 55.9 & 38.2 (+18.6)  & 34.6 & 56.6 & 40.2 (+16.6) \\
zh  & 45.0 & 61.9 & 47.8 (+14.7)  & 49.5 & 70.0 & 58.5 (+12.2) \\
km  & 26.2 & 28.7 & 27.9 (+5.9)  & 14.0 & 23.2 & 21.1 (+6.6) \\
\midrule
AVG & 50.1 & 65.9 & 58.5 (+7.0) & 38.4 & 70.7 & 59.1 (+7.7)\\
\bottomrule
\end{tabular}
\caption{Language-specific results on MKQA with single-passage mRAG setup. Numbers between brackets indicate the proportion of queries that are correctly answered but in the wrong language (i.e., not the query language).}
\label{tab:full_MKQA}
\end{table*}

\begin{table*}[!t]
\centering
\begin{tabular}{c|rrl|rrl}
\toprule
\multirow{2}{*}{\bf Language} & \multicolumn{3}{c|}{\bf Accuracy (Aya)} & \multicolumn{3}{c}{\bf Accuracy (Llama)} \\
& \textbf{Non.} & \textbf{In.} & \textbf{Out.} & \textbf{Non.} & \textbf{In.} & \textbf{Out.} \\
\midrule
en  & 47.9 & 70.4 & 32.7 (+30.8)  & 48.5 & 65.7 & 38.8 (+15.9) \\
ja  & 38.9 & 68.3 & 43.4 (+11.7)  & 19.9 & 46.9 & 23.4 (+6.6) \\
it  & 40.4 & 66.8 & 49.3 (+6.2)  & 26.9 & 51.9 & 34.7 (+6.1) \\
id  & 40.6 & 68.2 & 48.1 (+8.2)  & 27.2 & 46.7 & 30.4 (+6.0) \\
ko  & 12.4 & 25.7 & 16.6 (+11.7)  & 18.8 & 43.6 & 21.1 (+7.5) \\
nl  & 38.5 & 64.6 & 44.4 (+7.5)  & 29.2 & 50.2 & 31.0 (+5.7) \\
zh  & 40.7 & 65.2 & 47.8 (+10.8)  & 27.6 & 45.1 & 30.1 (+6.5) \\
vi  & 34.6 & 60.5 & 39.6 (+11.2)  & 26.1 & 45.4 & 28.1 (+8.6) \\
sv  & 22.4 & 55.6 & 32.7 (+8.0)  & 28.3 & 49.4 & 33.2 (+5.4) \\
pt  & 40.5 & 72.3 & 50.1 (+7.4)  & 27.3 & 53.7 & 33.7 (+6.1) \\
de  & 41.1 & 67.1 & 48.8 (+5.5)  & 32.0 & 53.4 & 37.6 (+5.1) \\
tr  & 36.0 & 65.7 & 39.0 (+13.7)  & 21.0 & 47.3 & 24.0 (+8.6) \\
ro  & 40.2 & 60.8 & 45.7 (+5.3)  & 26.5 & 44.2 & 32.5 (+4.6) \\
cs  & 34.2 & 57.2 & 39.0 (+7.7)  & 22.0 & 42.1 & 26.2 (+6.0) \\
ru  & 34.1 & 63.0 & 39.3 (+8.1)  & 21.0 & 43.3 & 23.3 (+6.5) \\
es  & 34.4 & 58.5 & 41.5 (+7.0)  & 25.7 & 48.4 & 34.3 (+5.4) \\
ms  & 35.8 & 64.1 & 44.0 (+8.7)  & 27.2 & 45.8 & 32.3 (+5.9) \\
pl  & 33.3 & 58.1 & 37.9 (+6.9)  & 20.6 & 39.1 & 24.8 (+6.3) \\
uk  & 32.8 & 59.4 & 37.4 (+6.7)  & 16.0 & 38.8 & 20.0 (+5.4) \\
fr  & 38.5 & 62.5 & 44.9 (+7.5)  & 23.8 & 50.7 & 31.5 (+5.5) \\
ar  & 29.2 & 59.9 & 36.0 (+9.4)  & 10.4 & 32.7 & 11.4 (+6.1) \\
fa  & 29.6 & 61.1 & 35.8 (+10.1)  & 15.2 & 43.6 & 16.9 (+8.7) \\
el  & 31.4 & 53.3 & 34.6 (+7.0)  & 15.5 & 35.6 & 17.7 (+8.1) \\
sr  & 13.4 & 38.4 & 18.9 (+7.7)  & 12.3 & 33.6 & 17.7 (+6.8) \\
he  & 30.9 & 60.6 & 36.6 (+9.3)  & 12.3 & 24.7 & 17.4 (+10.5) \\
hi  & 21.3 & 43.3 & 26.6 (+6.3)  & 17.9 & 39.0 & 23.4 (+1.5) \\
fil  & 23.4 & 44.9 & 30.2 (+8.7)  & 25.5 & 39.3 & 28.0 (+6.1) \\
lt  & 16.7 & 46.7 & 21.6 (+9.1)  & 14.8 & 35.9 & 19.0 (+4.9) \\
bn  & 5.1 & 23.9 & 8.2 (+5.4)  & 10.0 & 25.8 & 13.7 (+2.2) \\
ky  & 14.5 & 36.1 & 22.7 (+2.6)  & 13.9 & 27.4 & 16.6 (+5.9) \\
ha  & 15.0 & 43.9 & 24.3 (+16.0)  & 13.7 & 31.1 & 19.8 (+6.5) \\
te  & 4.8 & 15.5 & 6.2 (+2.9)  & 13.2 & 20.0 & 13.9 (+0.4) \\
sw  & 16.6 & 56.1 & 25.5 (+9.1)  & 20.1 & 34.7 & 25.8 (+4.7) \\
ig  & 15.5 & 34.9 & 20.8 (+13.5)  & 16.2 & 27.6 & 17.6 (+3.5) \\
si  & 6.1 & 13.3 & 4.4 (+3.3)  & 8.5 & 13.9 & 8.3 (+3.1) \\
ne  & 8.4 & 28.3 & 10.8 (+13.7)  & 9.2 & 27.7 & 10.2 (+14.4) \\
am  & 8.0 & 18.9 & 16.5 (+23.3)  & 8.5 & 10.3 & 5.8 (+0.3) \\
ny  & 21.5 & 44.2 & 29.6 (+13.8)  & 17.9 & 28.2 & 19.9 (+3.5) \\
mg  & 18.3 & 44.6 & 24.4 (+10.0)  & 20.1 & 40.4 & 22.1 (+4.6) \\
so  & 23.7 & 54.9 & 31.7 (+8.2)  & 19.9 & 40.0 & 22.7 (+4.6) \\
sn  & 27.5 & 60.2 & 31.9 (+15.7)  & 19.0 & 40.1 & 22.2 (+3.8) \\
yo  & 24.1 & 40.2 & 29.4 (+7.0)  & 20.2 & 32.9 & 24.7 (+1.1) \\
\midrule
AVG & 26.7 & 51.4 & 32.1 (+9.6) & 20.2 & 39.0 & 23.5 (+5.8) \\
\bottomrule
\end{tabular}
\caption{Language-specific results on GMMLU-Open with single-passage mRAG setup when the model is given no options and forced to output an open answer as the response. Numbers between brackets indicate the proportion of queries that are correctly answered but in the wrong language (i.e., not the query language).}
\label{tab:full_GMMLU_open_1}
\end{table*}

\begin{table*}[!t]
\centering
\begin{tabular}{c|rrl|rrl}
\toprule
\multirow{2}{*}{\bf Language} & \multicolumn{3}{c|}{\bf Accuracy (Gemma)} & \multicolumn{3}{c}{\bf Accuracy (Qwen)} \\
& \textbf{Non.} & \textbf{In.} & \textbf{Out.} & \textbf{Non.} & \textbf{In.} & \textbf{Out.} \\
\midrule
en  & 54.4 & 69.2 & 61.1 (+2.6)  & 56.1 & 73.2 & 63.4 (+2.2) \\
ja  & 43.0 & 67.5 & 47.7 (+5.1)  & 33.1 & 66.0 & 41.4 (+6.9) \\
it  & 47.2 & 65.7 & 54.0 (+4.7)  & 36.3 & 64.5 & 45.3 (+6.1) \\
id  & 46.8 & 66.8 & 50.6 (+7.4)  & 38.5 & 63.3 & 44.4 (+6.5) \\
ko  & 13.2 & 20.4 & 18.4 (+6.7)  & 23.7 & 43.7 & 28.3 (+7.4) \\
nl  & 45.0 & 63.2 & 46.6 (+6.8)  & 31.9 & 58.4 & 38.8 (+6.7) \\
zh  & 48.3 & 66.4 & 54.3 (+5.1)  & 52.3 & 67.2 & 57.7 (+4.1) \\
vi  & 38.5 & 58.1 & 39.6 (+9.8)  & 34.9 & 59.6 & 41.0 (+8.1) \\
sv  & 43.6 & 62.5 & 47.0 (+5.7)  & 28.4 & 60.0 & 35.6 (+7.2) \\
pt  & 47.3 & 68.9 & 51.8 (+6.3)  & 38.6 & 68.5 & 48.7 (+5.6) \\
de  & 47.2 & 66.8 & 54.2 (+4.2)  & 36.8 & 62.9 & 44.4 (+5.3) \\
tr  & 40.1 & 59.7 & 42.1 (+7.3)  & 24.2 & 56.2 & 32.2 (+10.1) \\
ro  & 43.2 & 58.3 & 46.6 (+5.2)  & 30.1 & 52.1 & 36.0 (+5.7) \\
cs  & 34.7 & 54.1 & 37.1 (+6.9)  & 23.2 & 47.9 & 28.5 (+5.7) \\
ru  & 38.1 & 55.7 & 38.4 (+8.2)  & 31.3 & 56.2 & 36.0 (+6.0) \\
es  & 39.8 & 57.7 & 45.5 (+6.2)  & 37.0 & 60.5 & 44.0 (+5.4) \\
ms  & 44.6 & 61.4 & 46.1 (+8.8)  & 32.5 & 57.4 & 39.7 (+7.2) \\
pl  & 36.3 & 54.0 & 38.9 (+6.8)  & 25.6 & 51.2 & 30.6 (+7.2) \\
uk  & 33.3 & 53.1 & 36.5 (+7.1)  & 18.3 & 45.8 & 23.8 (+5.4) \\
fr  & 39.5 & 62.1 & 47.7 (+5.8)  & 36.5 & 65.1 & 44.8 (+6.0) \\
ar  & 24.2 & 56.4 & 29.1 (+6.9)  & 22.4 & 57.8 & 29.2 (+7.3) \\
fa  & 31.4 & 61.2 & 36.3 (+7.7)  & 14.2 & 48.9 & 19.8 (+9.0) \\
el  & 29.4 & 48.1 & 29.2 (+7.4)  & 10.2 & 25.7 & 11.4 (+4.8) \\
sr  & 32.1 & 48.9 & 32.1 (+5.7)  & 17.5 & 43.4 & 23.8 (+5.6) \\
he  & 30.4 & 56.4 & 34.9 (+8.6)  & 18.8 & 53.3 & 26.7 (+7.9) \\
hi  & 35.1 & 52.2 & 35.7 (+2.9)  & 15.5 & 39.2 & 20.0 (+2.6) \\
fil  & 45.2 & 52.6 & 44.9 (+7.3)  & 30.0 & 45.9 & 33.4 (+9.7) \\
lt  & 31.4 & 51.0 & 29.9 (+8.0)  & 16.3 & 42.0 & 20.6 (+7.4) \\
bn  & 24.4 & 46.9 & 24.7 (+6.6)  & 11.0 & 35.5 & 15.0 (+5.2) \\
ky  & 28.2 & 43.5 & 30.9 (+3.3)  & 15.7 & 33.7 & 21.1 (+4.6) \\
ha  & 29.8 & 42.0 & 31.6 (+5.6)  & 19.2 & 41.4 & 24.4 (+8.9) \\
te  & 24.7 & 43.2 & 29.0 (+4.4)  & 7.7 & 17.7 & 6.9 (+1.6) \\
sw  & 35.6 & 48.7 & 37.4 (+6.1)  & 17.2 & 44.2 & 23.7 (+6.8) \\
ig  & 23.7 & 36.1 & 25.9 (+4.4)  & 17.4 & 35.7 & 24.1 (+7.7) \\
si  & 17.0 & 32.1 & 19.9 (+1.6)  & 9.5 & 15.7 & 8.8 (+1.0) \\
ne  & 24.1 & 38.4 & 25.8 (+11.7)  & 6.1 & 24.0 & 8.0 (+13.2) \\
am  & 15.6 & 24.6 & 16.9 (+2.2)  & 11.9 & 20.8 & 14.1 (+4.2) \\
ny  & 29.0 & 35.6 & 29.0 (+3.2)  & 18.0 & 29.9 & 24.0 (+4.7) \\
mg  & 26.7 & 33.0 & 24.2 (+4.2)  & 19.9 & 40.5 & 27.1 (+6.0) \\
so  & 29.2 & 40.6 & 28.6 (+4.1)  & 23.6 & 42.3 & 27.4 (+6.1) \\
sn  & 34.3 & 42.6 & 29.2 (+5.3)  & 20.2 & 37.4 & 26.4 (+5.2) \\
yo  & 21.2 & 35.3 & 25.4 (+2.5)  & 26.7 & 42.7 & 34.1 (+2.5) \\
\midrule
AVG & 34.4 & 51.5 & 37.0 (+5.9) & 24.7 & 47.6 & 30.3 (+6.1) \\
\bottomrule
\end{tabular}
\caption{Extension: Language-specific results on GMMLU-Open with single-passage mRAG setup when the model is given no options and forced to output an open answer as the response. Numbers between brackets indicate the proportion of queries that are correctly answered but in the wrong language (i.e., not the query language).}
\label{tab:full_GMMLU_open_2}
\end{table*}

\begin{table*}[!t]
\centering
\resizebox{0.99\linewidth}{!}{
\begin{tabular}{c|rrl|rrl|rrl|rrl}
\toprule
\multirow{2}{*}{\bf Language} & \multicolumn{3}{c|}{\bf Accuracy (Aya)} & \multicolumn{3}{c}{\bf Accuracy (Llama)} & \multicolumn{3}{c}{\bf Accuracy (Gemma)} & \multicolumn{3}{c}{\bf Accuracy (Qwen)}\\
& \textbf{Non.} & \textbf{In.} & \textbf{Out.} & \textbf{Non.} & \textbf{In.} & \textbf{Out.} & \textbf{Non.} & \textbf{In.} & \textbf{Out.} & \textbf{Non.} & \textbf{In.} & \textbf{Out.} \\
\midrule
en  & 70.2 & 77.3 & 75.5  & 69.7 & 76.6 & 72.6  & 80.5 & 83.3 & 81.0  & 81.6 & 84.3 & 82.8 \\
ja  & 61.5 & 71.1 & 71.4  & 44.3 & 58.1 & 57.9  & 69.9 & 77.2 & 75.9  & 66.2 & 74.0 & 71.4 \\
it  & 64.6 & 70.7 & 70.7  & 59.2 & 65.8 & 64.6  & 75.5 & 78.6 & 78.9  & 72.2 & 74.9 & 75.1 \\
id  & 61.8 & 69.7 & 68.9  & 50.7 & 57.6 & 57.9  & 72.5 & 76.7 & 75.4  & 69.8 & 74.5 & 73.2 \\
ko  & 58.8 & 64.6 & 65.8  & 39.3 & 47.1 & 47.7  & 66.7 & 72.6 & 71.9  & 64.4 & 70.7 & 69.2 \\
nl  & 61.7 & 68.8 & 67.9  & 54.3 & 63.6 & 61.5  & 73.2 & 75.9 & 74.9  & 71.0 & 73.6 & 74.3 \\
zh  & 60.3 & 66.2 & 68.8  & 52.3 & 61.5 & 59.2  & 71.0 & 74.0 & 75.3  & 73.2 & 72.2 & 72.5 \\
vi  & 55.5 & 62.2 & 62.4  & 48.5 & 57.8 & 56.0  & 65.8 & 71.1 & 70.9  & 66.9 & 70.7 & 69.8 \\
sv  & 52.7 & 65.3 & 65.2  & 47.3 & 60.3 & 58.5  & 72.3 & 76.4 & 76.4  & 67.0 & 73.6 & 72.9 \\
pt  & 70.1 & 75.5 & 75.0  & 35.1 & 60.7 & 53.3  & 78.2 & 83.2 & 81.6  & 78.6 & 83.0 & 82.1 \\
de  & 67.9 & 75.2 & 74.2  & 60.2 & 70.3 & 69.0  & 78.3 & 81.6 & 81.9  & 74.9 & 80.4 & 78.5 \\
tr  & 59.1 & 67.4 & 68.1  & 46.4 & 57.7 & 56.1  & 69.1 & 73.4 & 74.0  & 57.5 & 67.0 & 66.4 \\
ro  & 61.8 & 66.2 & 66.5  & 51.4 & 59.2 & 58.4  & 70.9 & 72.8 & 73.0  & 64.9 & 70.2 & 68.3 \\
cs  & 60.9 & 70.2 & 68.0  & 47.1 & 56.6 & 56.9  & 72.1 & 75.0 & 74.4  & 65.2 & 71.9 & 69.6 \\
ru  & 60.9 & 70.6 & 70.0  & 40.3 & 49.2 & 50.5  & 72.3 & 77.8 & 77.4  & 74.2 & 76.4 & 76.0 \\
es  & 63.9 & 69.5 & 69.7  & 57.5 & 65.8 & 64.0  & 73.5 & 76.3 & 76.4  & 72.4 & 76.9 & 76.0 \\
ms  & 56.3 & 65.4 & 64.4  & 47.0 & 54.4 & 54.9  & 70.1 & 73.2 & 72.7  & 66.2 & 72.3 & 70.8 \\
pl  & 59.4 & 68.9 & 67.7  & 47.9 & 55.2 & 51.5  & 71.7 & 75.1 & 74.9  & 67.2 & 71.6 & 70.5 \\
uk  & 58.9 & 68.4 & 67.2  & 29.1 & 34.8 & 37.0  & 70.5 & 75.3 & 74.9  & 64.8 & 72.4 & 70.6 \\
fr  & 69.5 & 78.1 & 77.1  & 62.3 & 67.2 & 67.0  & 80.4 & 84.1 & 82.8  & 77.9 & 82.0 & 81.0 \\
ar  & 60.7 & 74.3 & 73.4  & 41.2 & 61.4 & 58.0  & 66.1 & 77.5 & 76.7  & 63.2 & 75.9 & 73.2 \\
fa  & 58.4 & 68.9 & 69.2  & 38.8 & 53.2 & 53.3  & 69.3 & 78.6 & 78.1  & 58.3 & 71.3 & 68.4 \\
el  & 58.1 & 65.0 & 65.2  & 27.8 & 40.8 & 45.1  & 64.7 & 72.4 & 71.9  & 47.3 & 61.4 & 58.0 \\
sr  & 41.0 & 58.2 & 55.8  & 14.5 & 8.8 & 21.3  & 65.1 & 70.6 & 70.5  & 58.4 & 67.7 & 66.5 \\
he  & 54.3 & 64.0 & 62.7  & 19.4 & 18.8 & 24.5  & 59.7 & 69.6 & 69.8  & 28.1 & 35.4 & 40.1 \\
hi  & 51.0 & 58.3 & 60.8  & 39.1 & 49.1 & 49.7  & 64.4 & 66.8 & 68.6  & 49.0 & 61.2 & 59.1 \\
fil  & 41.9 & 48.0 & 50.7  & 37.2 & 39.1 & 40.7  & 66.6 & 68.7 & 70.0  & 58.0 & 60.4 & 59.1 \\
lt  & 39.9 & 55.0 & 50.9  & 34.0 & 48.5 & 46.6  & 65.4 & 71.4 & 70.7  & 46.6 & 61.3 & 56.9 \\
bn  & 26.7 & 48.9 & 46.6  & 30.2 & 45.4 & 49.1  & 61.0 & 67.6 & 67.8  & 52.3 & 62.1 & 61.5 \\
ky  & 26.4 & 37.1 & 41.9  & 20.4 & 24.3 & 28.5  & 48.2 & 53.5 & 54.1  & 38.7 & 49.9 & 48.2 \\
ha  & 29.4 & 38.3 & 33.7  & 26.6 & 30.8 & 31.8  & 35.4 & 32.8 & 38.9  & 25.6 & 35.5 & 30.4 \\
te  & 11.9 & 21.6 & 28.7  & 29.8 & 44.2 & 42.4  & 56.8 & 61.2 & 62.6  & 25.9 & 33.4 & 32.2 \\
sw  & 28.5 & 39.8 & 35.7  & 32.0 & 37.0 & 40.9  & 55.2 & 57.4 & 60.1  & 27.2 & 37.2 & 32.8 \\
ig  & 28.0 & 35.8 & 33.3  & 25.1 & 30.0 & 28.7  & 33.2 & 40.2 & 39.6  & 22.7 & 31.0 & 29.7 \\
si  & 7.1 & 12.6 & 12.5  & 22.6 & 32.1 & 26.7  & 35.9 & 40.6 & 49.9  & 10.7 & 15.7 & 15.8 \\
ne  & 35.6 & 40.7 & 44.1  & 26.5 & 25.7 & 31.9  & 56.2 & 55.9 & 59.6  & 36.7 & 38.7 & 40.7 \\
am  & 3.2 & 5.9 & 17.3  & 15.4 & 5.5 & 8.2  & 36.1 & 45.4 & 44.0  & 10.3 & 25.0 & 17.9 \\
ny  & 22.2 & 24.6 & 23.6  & 21.5 & 21.5 & 26.4  & 37.8 & 46.0 & 43.8  & 19.3 & 28.1 & 25.5 \\
mg  & 22.5 & 32.5 & 27.4  & 19.3 & 22.1 & 23.7  & 36.5 & 37.0 & 36.2  & 21.1 & 28.2 & 26.2 \\
so  & 28.1 & 38.3 & 34.7  & 24.1 & 27.0 & 26.0  & 26.4 & 35.9 & 36.7  & 25.2 & 36.4 & 31.6 \\
sn  & 21.2 & 22.6 & 24.9  & 23.0 & 23.0 & 27.1  & 39.4 & 44.2 & 46.1  & 19.8 & 21.0 & 23.5 \\
yo  & 21.7 & 19.8 & 25.7  & 20.7 & 25.0 & 25.0  & 32.7 & 38.7 & 37.1  & 17.0 & 19.2 & 21.5 \\
\midrule
AVG & 46.5 & 54.8 & 54.8 & 37.6 & 45.1 & 45.5 & 61.1 & 65.8 & 66.1 & 51.4 & 58.3 & 56.9 \\
\bottomrule
\end{tabular}
}
\caption{
Language-specific results on GMMLU-Choice with single-passage mRAG setup when the model is given options and the answer accuracy is evaluated by whether the model outputs the correct option letter. This setup eliminates the effect of generation language on the performance evaluation.
}
\label{tab:full_GMMLU_choice}
\end{table*}

\begin{table*}[!t]
\centering
\resizebox{0.99\linewidth}{!}{
\begin{tabular}{l|cccccccccccccc}
\toprule
\rowcolor{black!10} \multicolumn{15}{c}{\bf Accuracy (Aya)}\\
\midrule
Setups & en & ja & it & id & ko & nl & zh & vi & sv & pt & de & tr & ro & cs \\
\midrule
No Ctx & 70.3 & 61.4 & 64.8 & 61.9 & 58.8 & 61.7 & 60.1 & 55.6 & 53.1 & 70.5 & 68.1 & 59.5 & 61.9 & 61.2 \\
\midrule
1o & 75.2 & 71.3 & 71.0 & 69.1 & 65.8 & 68.3 & 67.1 & 63.4 & 65.6 & 75.0 & 74.7 & 67.4 & 66.3 & 68.4 \\
1o3i & 72.9 & 68.1 & 68.8 & 64.8 & 62.4 & 63.4 & 62.8 & 57.7 & 61.3 & 64.5 & 70.2 & 64.0 & 62.4 & 65.6 \\
1o3o & 72.2 & 68.0 & 68.4 & 63.7 & 62.9 & 64.6 & 62.8 & 56.6 & 61.6 & 68.7 & 70.3 & 63.0 & 61.1 & 64.6 \\
\midrule
3o & 78.3 & 74.5 & 73.6 & 72.2 & 69.2 & 70.4 & 70.7 & 63.3 & 68.4 & 73.9 & 77.5 & 69.9 & 67.0 & 71.2 \\
3o1i & 77.2 & 73.5 & 73.5 & 70.9 & 68.7 & 69.3 & 69.0 & 62.0 & 67.7 & 71.2 & 76.0 & 68.7 & 66.8 & 70.8 \\
3o1o & 77.4 & 73.9 & 73.5 & 71.0 & 69.3 & 69.6 & 69.5 & 61.6 & 67.3 & 74.3 & 76.5 & 69.3 & 67.0 & 71.0 \\
\bottomrule
Setups & ru & es & ms & pl & uk & fr & ar & fa & el & sr & he & hi & fil & lt \\
\midrule
No Ctx & 60.9 & 64.0 & 56.4 & 59.6 & 59.1 & 69.6 & 60.9 & 58.3 & 58.2 & 41.3 & 54.3 & 51.0 & 41.6 & 39.7 \\
\midrule
1o & 70.3 & 70.3 & 64.2 & 68.6 & 68.0 & 76.8 & 71.7 & 68.9 & 65.4 & 56.6 & 63.7 & 60.0 & 52.1 & 51.7 \\
1o3i & 65.1 & 66.6 & 59.3 & 65.6 & 64.4 & 73.6 & 68.0 & 64.6 & 61.9 & 52.5 & 60.6 & 57.7 & 44.3 & 46.3 \\
1o3o & 64.9 & 65.5 & 60.0 & 64.8 & 63.8 & 72.3 & 67.0 & 64.7 & 62.1 & 53.2 & 62.0 & 57.8 & 49.4 & 47.3 \\
\midrule
3o & 71.9 & 72.3 & 66.8 & 70.4 & 69.8 & 79.7 & 75.8 & 71.8 & 68.7 & 60.3 & 66.9 & 65.0 & 53.5 & 53.1 \\
3o1i & 70.8 & 71.0 & 65.6 & 70.2 & 69.4 & 77.7 & 74.7 & 71.4 & 67.9 & 58.8 & 65.7 & 64.0 & 51.9 & 52.8 \\
3o1o & 71.8 & 71.4 & 66.2 & 70.2 & 68.9 & 78.3 & 74.3 & 70.7 & 68.2 & 59.4 & 66.0 & 64.3 & 52.4 & 52.5 \\
\bottomrule
Setups & bn & ky & ha & te & sw & ig & si & ne & am & ny & mg & so & sn & yo \\
\midrule
No Ctx & 26.7 & 26.5 & 28.2 & 12.2 & 28.3 & 28.1 & 6.6 & 35.3 & 3.2 & 22.3 & 23.5 & 28.1 & 22.4 & 23.5 \\
\midrule
1o & 47.1 & 39.8 & 33.9 & 28.9 & 37.0 & 34.2 & 12.7 & 44.5 & 17.3 & 24.8 & 28.5 & 35.9 & 23.5 & 24.5 \\
1o3i & 42.6 & 34.2 & 33.1 & 14.7 & 31.9 & 30.1 & 12.1 & 41.1 & 9.0 & 21.2 & 25.2 & 33.9 & 23.8 & 23.0 \\
1o3o & 46.1 & 37.5 & 32.7 & 17.7 & 33.5 & 31.1 & 11.7 & 43.4 & 10.1 & 20.9 & 26.4 & 33.0 & 23.5 & 26.2 \\
\midrule
3o & 53.5 & 44.4 & 34.3 & 22.1 & 38.0 & 33.0 & 12.0 & 47.2 & 11.7 & 21.9 & 27.6 & 38.2 & 23.8 & 25.5 \\
3o1i & 51.4 & 40.2 & 35.8 & 20.4 & 37.0 & 34.0 & 13.3 & 44.5 & 7.8 & 22.6 & 30.1 & 36.0 & 26.9 & 25.8 \\
3o1o & 53.5 & 41.9 & 34.5 & 22.0 & 39.0 & 33.7 & 11.7 & 47.2 & 10.3 & 20.5 & 28.4 & 37.0 & 26.0 & 28.0 \\
\bottomrule
\end{tabular}
}
\caption{Full performance on GMMLU-Choice with multiple-passage mRAG setup.}
\label{tab:multi_mirage_full_1}
\end{table*}

\begin{table*}[!t]
\centering
\resizebox{0.99\linewidth}{!}{
\begin{tabular}{l|cccccccccccccc}
\toprule
\rowcolor{black!10} \multicolumn{15}{c}{\bf Accuracy (Llama)}\\
\midrule
Setups & en & ja & it & id & ko & nl & zh & vi & sv & pt & de & tr & ro & cs \\
\midrule
No Ctx & 70.0 & 43.6 & 58.9 & 51.1 & 39.4 & 54.7 & 52.2 & 48.7 & 47.8 & 34.2 & 60.3 & 46.4 & 51.4 & 48.1 \\
\midrule
1o & 73.0 & 57.5 & 64.5 & 58.7 & 48.1 & 62.2 & 58.9 & 56.6 & 59.2 & 51.0 & 69.3 & 56.6 & 58.4 & 57.1 \\
1o3i & 72.5 & 56.7 & 63.5 & 52.8 & 45.4 & 61.0 & 57.7 & 53.9 & 48.6 & 48.7 & 65.5 & 52.5 & 56.5 & 54.3 \\
1o3o & 71.3 & 55.2 & 60.1 & 53.7 & 45.5 & 60.6 & 56.5 & 54.9 & 53.1 & 49.3 & 66.3 & 53.3 & 56.8 & 54.3 \\
\midrule
3o & 75.5 & 63.9 & 68.3 & 59.4 & 52.2 & 65.4 & 63.6 & 61.1 & 59.0 & 53.0 & 73.6 & 60.0 & 62.8 & 60.3 \\
3o1i & 75.7 & 62.7 & 67.4 & 56.8 & 50.3 & 64.8 & 62.7 & 59.2 & 55.7 & 49.6 & 73.1 & 58.4 & 61.4 & 58.4 \\
3o1o & 75.2 & 63.1 & 66.6 & 58.1 & 51.8 & 65.3 & 62.4 & 60.4 & 57.8 & 53.1 & 73.3 & 58.5 & 61.3 & 59.5 \\
\bottomrule
Setups & ru & es & ms & pl & uk & fr & ar & fa & el & sr & he & hi & fil & lt \\
\midrule
No Ctx & 41.0 & 57.2 & 47.3 & 48.2 & 29.1 & 62.9 & 41.2 & 38.6 & 27.3 & 14.5 & 20.6 & 39.2 & 37.6 & 34.5 \\
\midrule
1o & 49.7 & 64.2 & 55.0 & 53.0 & 36.7 & 67.3 & 59.0 & 52.5 & 46.4 & 22.6 & 25.8 & 48.4 & 41.5 & 45.7 \\
1o3i & 33.1 & 64.6 & 49.4 & 36.7 & 16.5 & 65.4 & 58.3 & 46.1 & 21.9 & 5.1 & 28.2 & 46.1 & 39.5 & 36.9 \\
1o3o & 41.9 & 62.1 & 52.3 & 46.6 & 19.7 & 65.0 & 57.8 & 45.0 & 30.3 & 11.6 & 26.9 & 48.1 & 38.4 & 42.7 \\
\midrule
3o & 49.1 & 68.6 & 58.7 & 52.7 & 25.0 & 71.5 & 67.3 & 52.6 & 39.9 & 14.3 & 35.4 & 54.7 & 42.6 & 49.2 \\
3o1i & 41.1 & 67.7 & 56.0 & 47.1 & 18.4 & 71.9 & 65.8 & 51.7 & 30.0 & 6.7 & 34.8 & 51.8 & 43.3 & 45.6 \\
3o1o & 47.3 & 67.6 & 56.6 & 52.5 & 21.6 & 70.9 & 65.5 & 50.2 & 33.8 & 12.5 & 32.0 & 53.6 & 42.6 & 47.9 \\
\bottomrule
Setups & bn & ky & ha & te & sw & ig & si & ne & am & ny & mg & so & sn & yo \\
\midrule
No Ctx & 30.3 & 21.1 & 26.2 & 28.7 & 32.3 & 25.0 & 22.1 & 26.8 & 15.9 & 22.3 & 19.8 & 23.9 & 24.8 & 21.4 \\
\midrule
1o & 48.2 & 28.7 & 30.6 & 42.6 & 41.7 & 29.0 & 25.1 & 30.2 & 8.1 & 26.0 & 24.0 & 28.5 & 26.2 & 24.4 \\
1o3i & 49.3 & 23.3 & 28.6 & 43.1 & 42.3 & 28.2 & 25.8 & 21.4 & 7.8 & 22.0 & 21.0 & 24.9 & 20.7 & 29.5 \\
1o3o & 48.7 & 20.3 & 30.6 & 40.0 & 45.2 & 28.0 & 26.9 & 24.5 & 8.5 & 27.6 & 22.5 & 27.1 & 28.3 & 26.1 \\
\midrule
3o & 55.9 & 21.5 & 34.1 & 47.3 & 49.4 & 28.7 & 28.9 & 29.3 & 8.6 & 26.9 & 25.5 & 28.8 & 29.2 & 28.3 \\
3o1i & 56.4 & 19.6 & 32.4 & 49.5 & 48.1 & 26.9 & 30.2 & 27.5 & 9.2 & 24.1 & 22.6 & 28.2 & 26.2 & 30.3 \\
3o1o & 56.0 & 21.8 & 33.9 & 47.7 & 50.4 & 28.7 & 28.7 & 27.5 & 9.1 & 26.8 & 23.3 & 27.9 & 29.9 & 26.5 \\
\bottomrule
\end{tabular}
}
\caption{Extension: Full performance on GMMLU-Choice with multiple-passage mRAG setup.}
\label{tab:multi_mirage_full_2}
\end{table*}

\begin{table*}[!t]
\centering
\resizebox{0.99\linewidth}{!}{
\begin{tabular}{l|cccccccccccccc}
\toprule
\rowcolor{black!10} \multicolumn{15}{c}{\bf Accuracy (Gemma}\\
\midrule
Setups & en & ja & it & id & ko & nl & zh & vi & sv & pt & de & tr & ro & cs \\
\midrule
No Ctx & 80.6 & 69.8 & 75.6 & 72.8 & 66.7 & 73.1 & 71.0 & 65.9 & 72.3 & 78.3 & 78.4 & 69.0 & 71.0 & 72.3 \\
\midrule
1o & 81.2 & 76.4 & 78.6 & 75.9 & 72.1 & 74.6 & 75.1 & 71.7 & 76.1 & 82.1 & 81.7 & 74.3 & 73.8 & 74.7 \\
1o3i & 80.4 & 76.3 & 78.2 & 74.8 & 71.4 & 74.8 & 73.2 & 70.1 & 75.1 & 78.2 & 80.1 & 72.9 & 72.6 & 74.3 \\
1o3o & 80.2 & 74.4 & 77.1 & 74.2 & 71.4 & 73.4 & 72.6 & 68.7 & 73.9 & 74.7 & 80.4 & 71.5 & 71.7 & 73.1 \\
\midrule
3o & 83.1 & 78.6 & 80.8 & 78.6 & 75.9 & 76.8 & 77.1 & 73.7 & 77.8 & 81.7 & 83.7 & 76.4 & 74.7 & 77.7 \\
3o1i & 82.9 & 78.9 & 81.2 & 78.3 & 75.8 & 77.4 & 76.9 & 72.4 & 77.6 & 80.5 & 83.1 & 76.2 & 75.4 & 77.3 \\
3o1o & 82.9 & 78.7 & 80.8 & 78.4 & 76.1 & 76.7 & 76.3 & 73.1 & 77.5 & 79.5 & 84.1 & 75.8 & 74.5 & 76.9 \\
\bottomrule
Setups & ru & es & ms & pl & uk & fr & ar & fa & el & sr & he & hi & fil & lt \\
\midrule
No Ctx & 72.2 & 73.5 & 70.4 & 72.0 & 70.7 & 80.5 & 66.1 & 69.3 & 64.9 & 65.3 & 59.5 & 64.4 & 66.9 & 65.6 \\
\midrule
1o & 77.6 & 76.8 & 72.5 & 75.5 & 75.4 & 83.2 & 75.8 & 77.4 & 72.4 & 70.3 & 70.2 & 68.2 & 69.8 & 70.2 \\
1o3i & 75.6 & 76.2 & 71.1 & 74.8 & 73.1 & 81.9 & 74.8 & 75.3 & 71.3 & 70.5 & 68.7 & 67.5 & 69.5 & 69.3 \\
1o3o & 75.5 & 75.0 & 70.8 & 73.4 & 73.2 & 82.0 & 72.6 & 75.2 & 71.0 & 68.7 & 69.1 & 67.3 & 67.3 & 66.5 \\
\midrule
3o & 80.1 & 78.9 & 74.8 & 76.9 & 77.8 & 85.8 & 79.2 & 79.9 & 75.1 & 73.6 & 73.8 & 71.9 & 71.7 & 72.2 \\
3o1i & 79.3 & 78.4 & 74.7 & 76.8 & 76.8 & 85.2 & 78.9 & 80.0 & 74.1 & 74.0 & 73.7 & 71.4 & 71.7 & 72.7 \\
3o1o & 79.5 & 78.4 & 74.1 & 76.7 & 76.5 & 85.1 & 78.5 & 79.9 & 74.3 & 72.9 & 72.8 & 71.0 & 71.5 & 71.3 \\
\bottomrule
Setups & bn & ky & ha & te & sw & ig & si & ne & am & ny & mg & so & sn & yo \\
\midrule
No Ctx & 61.1 & 48.1 & 34.6 & 56.8 & 55.2 & 33.3 & 36.2 & 56.2 & 35.5 & 38.1 & 36.5 & 25.7 & 39.8 & 32.5 \\
\midrule
1o & 69.3 & 53.3 & 39.0 & 62.2 & 59.4 & 41.5 & 51.0 & 58.0 & 45.1 & 44.2 & 35.9 & 34.6 & 45.1 & 38.5 \\
1o3i & 68.3 & 52.3 & 35.4 & 60.1 & 58.8 & 35.7 & 49.5 & 58.7 & 44.1 & 45.4 & 30.7 & 42.7 & 44.2 & 37.9 \\
1o3o & 68.2 & 52.4 & 44.1 & 60.4 & 56.9 & 37.1 & 53.4 & 56.1 & 47.6 & 42.3 & 32.7 & 44.2 & 45.7 & 37.7 \\
\midrule
3o & 73.3 & 59.8 & 45.5 & 66.7 & 61.8 & 40.8 & 59.2 & 59.9 & 48.9 & 45.8 & 34.3 & 49.3 & 49.0 & 41.2 \\
3o1i & 72.2 & 58.0 & 41.1 & 66.0 & 61.3 & 39.0 & 53.9 & 60.3 & 48.4 & 47.7 & 34.4 & 49.1 & 48.1 & 40.5 \\
3o1o & 72.8 & 58.2 & 45.4 & 65.8 & 62.2 & 40.1 & 57.2 & 59.5 & 49.6 & 44.1 & 34.4 & 48.9 & 47.8 & 38.6 \\
\bottomrule
\end{tabular}
}
\caption{Extension: Full performance on GMMLU-Choice with multiple-passage mRAG setup.}
\label{tab:multi_mirage_full_3}
\end{table*}

\begin{table*}[!t]
\centering
\resizebox{0.99\linewidth}{!}{
\begin{tabular}{l|cccccccccccccc}
\toprule
\rowcolor{black!10} \multicolumn{15}{c}{\bf Accuracy (Qwen)}\\
\midrule
Setups & en & ja & it & id & ko & nl & zh & vi & sv & pt & de & tr & ro & cs \\
\midrule
No Ctx & 81.6 & 66.2 & 72.4 & 70.1 & 64.4 & 71.2 & 73.7 & 66.8 & 67.0 & 78.8 & 75.0 & 57.7 & 65.0 & 65.4 \\
\midrule
1o & 83.0 & 71.2 & 75.4 & 73.6 & 69.9 & 73.7 & 72.4 & 69.8 & 72.7 & 81.9 & 77.8 & 66.8 & 69.0 & 70.1 \\
1o3i & 80.8 & 70.3 & 74.4 & 72.0 & 68.8 & 70.9 & 69.0 & 65.3 & 71.0 & 79.5 & 76.1 & 65.0 & 65.3 & 68.8 \\
1o3o & 80.2 & 69.3 & 75.9 & 71.5 & 66.9 & 71.5 & 68.6 & 65.2 & 71.2 & 79.2 & 75.4 & 64.9 & 64.4 & 67.3 \\
\midrule
3o & 83.9 & 75.4 & 78.2 & 75.8 & 72.6 & 75.5 & 74.1 & 70.8 & 76.1 & 83.5 & 81.3 & 71.2 & 70.4 & 73.0 \\
3o1i & 83.7 & 74.7 & 77.7 & 75.3 & 71.9 & 74.9 & 74.1 & 69.9 & 75.9 & 83.8 & 81.3 & 69.2 & 69.3 & 72.9 \\
3o1o & 83.9 & 74.9 & 78.2 & 75.9 & 71.8 & 75.7 & 73.0 & 70.7 & 75.4 & 83.4 & 80.6 & 69.7 & 69.8 & 72.7 \\
\bottomrule
Setups & ru & es & ms & pl & uk & fr & ar & fa & el & sr & he & hi & fil & lt \\
\midrule
No Ctx & 74.3 & 72.5 & 66.5 & 67.3 & 65.0 & 77.9 & 63.2 & 58.2 & 47.3 & 58.3 & 28.6 & 48.9 & 58.0 & 46.7 \\
\midrule
1o & 76.1 & 75.7 & 70.2 & 71.8 & 70.8 & 81.4 & 72.6 & 68.4 & 58.1 & 66.6 & 40.5 & 57.9 & 59.4 & 56.1 \\
1o3i & 72.0 & 74.4 & 67.0 & 69.2 & 68.4 & 79.1 & 69.8 & 63.9 & 57.0 & 64.0 & 41.2 & 55.5 & 57.2 & 54.3 \\
1o3o & 72.2 & 74.6 & 67.9 & 67.9 & 68.5 & 78.6 & 67.6 & 62.9 & 54.8 & 63.9 & 40.4 & 56.9 & 56.5 & 54.3 \\
\midrule
3o & 78.5 & 78.1 & 73.3 & 72.8 & 73.2 & 83.5 & 75.3 & 73.4 & 62.4 & 69.6 & 44.2 & 63.5 & 62.4 & 60.0 \\
3o1i & 77.5 & 77.0 & 72.0 & 73.4 & 71.9 & 82.9 & 74.0 & 71.0 & 59.7 & 69.1 & 41.5 & 62.4 & 60.6 & 58.9 \\
3o1o & 77.2 & 77.4 & 72.5 & 73.1 & 72.6 & 83.4 & 73.4 & 70.6 & 60.4 & 68.5 & 42.4 & 62.3 & 61.1 & 59.5 \\
\bottomrule
Setups & bn & ky & ha & te & sw & ig & si & ne & am & ny & mg & so & sn & yo \\
\midrule
No Ctx & 52.2 & 38.9 & 26.0 & 25.6 & 27.3 & 22.5 & 9.8 & 36.5 & 9.4 & 18.9 & 21.3 & 26.2 & 19.7 & 15.8 \\
\midrule
1o & 61.6 & 48.0 & 30.8 & 33.2 & 33.2 & 29.0 & 15.5 & 41.0 & 18.6 & 27.4 & 25.4 & 31.5 & 24.5 & 20.5 \\
1o3i & 52.4 & 45.0 & 29.4 & 23.9 & 30.4 & 26.5 & 16.1 & 40.1 & 16.0 & 26.5 & 22.0 & 26.8 & 22.9 & 21.7 \\
1o3o & 49.7 & 43.6 & 31.6 & 20.2 & 29.7 & 28.8 & 19.4 & 42.5 & 18.1 & 28.9 & 24.9 & 29.0 & 27.4 & 22.7 \\
\midrule
3o & 60.8 & 50.1 & 33.3 & 28.5 & 35.0 & 31.5 & 20.7 & 48.9 & 20.7 & 29.5 & 26.2 & 31.4 & 27.2 & 24.7 \\
3o1i & 55.2 & 46.6 & 32.5 & 23.1 & 33.3 & 29.3 & 20.8 & 45.9 & 22.1 & 29.5 & 25.8 & 30.0 & 26.9 & 25.3 \\
3o1o & 55.0 & 47.2 & 33.1 & 25.0 & 34.2 & 29.5 & 21.1 & 48.1 & 20.7 & 29.0 & 25.2 & 30.7 & 27.6 & 23.5 \\
\bottomrule

\end{tabular}
}
\caption{Extension: Full performance on GMMLU-Choice with multiple-passage mRAG setup.}
\label{tab:multi_mirage_full_4}
\end{table*}

\end{document}